\documentclass[10pt,twocolumn,letterpaper]{article}

\usepackage{iccv}
\usepackage{times}
\usepackage{epsfig}
\usepackage{graphicx}
\usepackage{amsmath}
\usepackage{amssymb}

\usepackage{bbm}
\usepackage[ruled,vlined,noend]{algorithm2e}

\usepackage{rotating}
\usepackage{commath}
\usepackage{algorithmic}
\usepackage{adjustbox}
\usepackage{booktabs}

\usepackage{caption}

\usepackage[accsupp]{axessibility} 

\usepackage[breaklinks=true,bookmarks=false]{hyperref}

\iccvfinalcopy 

\def\iccvPaperID{****} 

\ificcvfinal\fi

\begin{document}

\title{Interpretable Visual Reasoning via Induced Symbolic Space}

\author{Zhonghao Wang$^{1}$, Kai Wang$^{3}$, Mo Yu$^{2}$\thanks{Correspondence to H. Shi and M. Yu.},  \\
Jinjun Xiong$^{2}$, Wen-mei Hwu$^{1}$, Mark Hasegawa-Johnson$^{1}$, Humphrey Shi$^{3,1,4}$\footnotemark[1]\\
\\

{\small $^1$UIUC, $^2$MIT-IBM Watson AI Lab \& IBM Research, $^3$U of Oregon, $^4$Picsart AI Research (PAIR)}}

\maketitle
\ificcvfinal\thispagestyle{empty}\fi

\begin{abstract}
We study the problem of concept induction in visual reasoning, i.e., identifying concepts and their hierarchical relationships from question-answer pairs associated with images; and achieve an interpretable model via working on the induced symbolic concept space.
To this end, we first design a new framework named object-centric compositional attention model (OCCAM) to perform the visual reasoning task with object-level visual features. Then, we come up with a method to induce concepts of objects and relations using clues from the attention patterns between objects' visual features and question words.
Finally, we achieve a higher level of interpretability by imposing OCCAM on the objects represented in the induced symbolic concept space.
%
Experiments on the CLEVR and GQA datasets demonstrate: 1) our OCCAM achieves a new state of the art without human-annotated functional programs; 2) our induced concepts are both accurate and sufficient as OCCAM achieves an on-par performance on objects represented either in visual features or in the induced symbolic concept space.
\end{abstract}

\vspace{-3mm}
\section{Introduction}

\begin{figure}[t]
\centering
\includegraphics[width=0.49\textwidth]{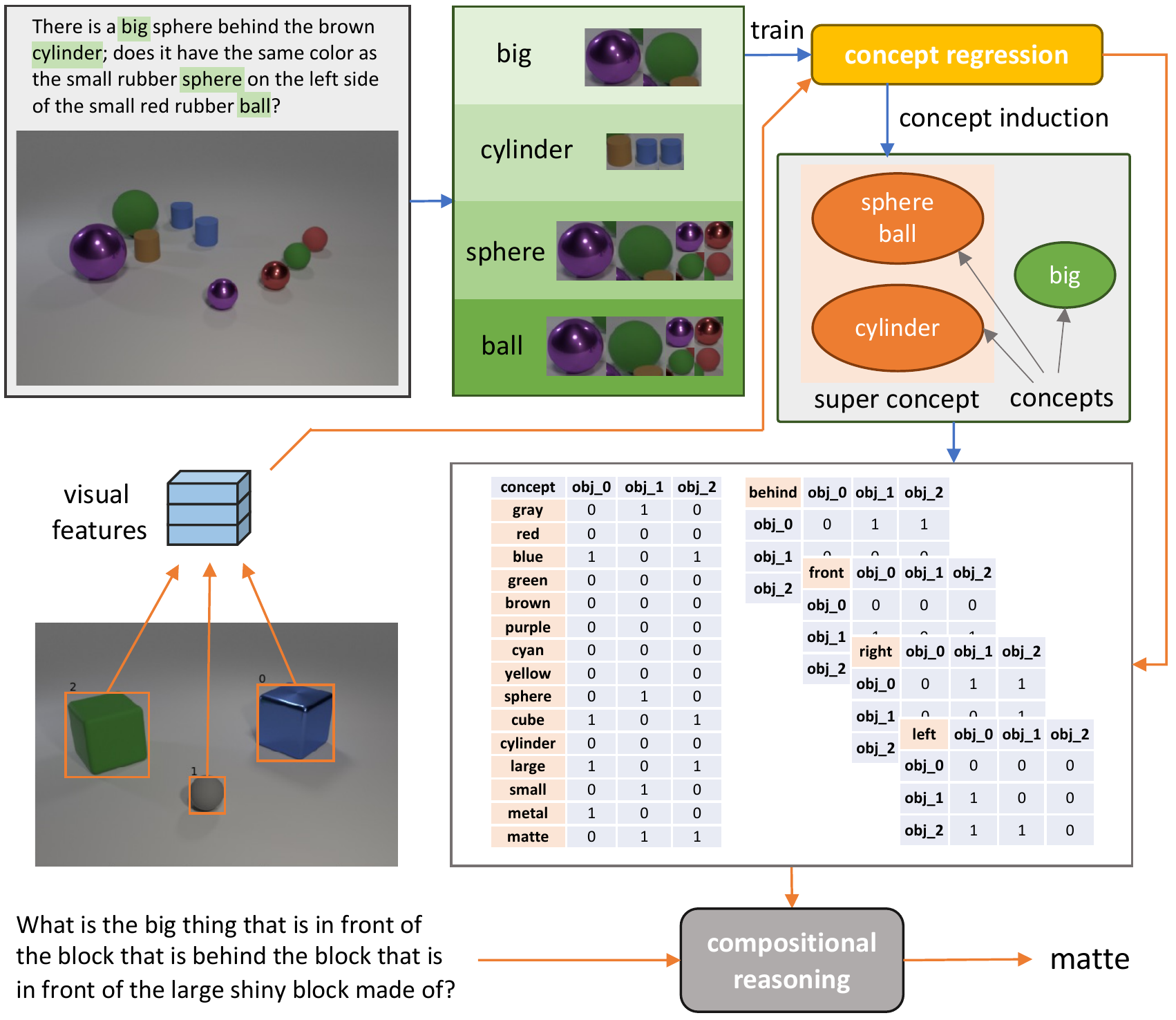}
\vspace{-3mm}
\caption{\small{Illustration of our framework. Our model induces the concepts and super concepts with the attention correlation between the objects and question words in image-question pairs as the paths shown in blue arrows. Then, it answers a question about an image via compositional reasoning on the induced symbolic representations of objects and object relations, shown as the orange paths.}}
\label{teaser}
\vspace{-6mm}
\end{figure}

Recent advances in Visual Question Answering (VQA)~\cite{anderson2018bottom, perez2018film,hudson2018compositional, andreas2016neural, hu2017learning, johnson2017inferring, hu2018explainable, yi2018neural, mao2018neuro} usually rely on carefully designed neural attention models over images, and rely on pre-defined lists of concepts to enhance the compositional reasoning ability of the attention modules.
Human prior knowledge plays an essential role in the success of the model design.

We focus on a less-studied problem in this field -- given only question-answer pairs and images, induce the visual concepts that are sufficient for completing the visual reasoning tasks. 
By sufficiency, we hope to maintain the predictive accuracy for VQA when using the induced concepts in place of the original visual features.
We consider concepts that are important for visual reasoning, including properties of objects (e.g., \emph{red}, \emph{cube}) and relations between objects (e.g., \emph{left}, \emph{front}).
The aforementioned scope and sufficiency criterion require accurately associating the induced symbols of concepts to both visual features and words, so that each new instance of question-image pair can be transformed into the induced concept space for further computations.
Additionally, it is necessary to identify super concepts, i.e., hypernyms of concept subsets (e.g., \emph{shape}). The concepts inside a super concept are exclusive, so that the system knows each object can only possess one value in each subset.
This introduces structural information to the concept space (multiple one-hot vectors for each visual object) and further guarantees the accuracy of the aforementioned transformation.

The value of the study has two folds.
First, our proposed problem aims to identify visual concepts, their argument patterns (properties or relations) and their hierarchy (super concepts) \emph{without} using any concept-level supervision. Solving the problem frees both the efforts of human annotations and human designs of concept schema required in previous visual reasoning works. At the same time, the problem is technically more challenging compared to the related existing problem like unsupervised or weakly-supervised visual grounding~\cite{yeh2018unsupervised}.
Second, by constraining the visual reasoning models to work over the induced concepts, the ability of concept induction improves the interpretability of visual reasoning models.
Unlike previous interpretable visual reasoning models that rely on human-written rules to associate neural modules with given concept definitions~\cite{hu2018explainable,mao2018neuro,shi2019explainable}, our method resolves the concept definitions and associations interpretability automatically in the learning process, without the need of trading off for hand-crafted model designs.


We achieve the proposed task in three steps.
First, we propose a new model architecture, object-centric compositional attention model (OCCAM), that performs object-level visual reasoning instead of pixel-level by extracting object-level visual features with ResNet \cite{He_2016_CVPR} and pooling the features according to each object's bounding box.
The object-level reasoning not only improves over the state-of-the-arts, but also provides a higher-level interpretability for concept association and identification.
Second, we benefit from the trained OCCAM's attention values over objects to create classifiers mapping visual objects to words; then derive the concepts and super concepts from the object-word cooccurrence matrices as shown in Figure \ref{teaser}.
Finally, our concept-based visual reasoning framework predicts the concepts of objects and object relations; then performs compositional reasoning using the predicted symbolic concept embeddings instead of the original visual features.



Experiments on the CLEVR and GQA datasets confirm that our overall approach improves the interpretability of neural visual reasoning, and maintains the predictive accuracy: (1) our OCCAM improves over the previous state-of-the-art models that do not use external training data; (2) our induced concepts and concept hierarchy are accurate in human study; and (3) our induced concepts are sufficient for visual reasoning -- replacing visual features with concepts leads to only $\sim$1\% performance drop.

\begin{figure*}[t]
\centering
\includegraphics[width=1\textwidth]{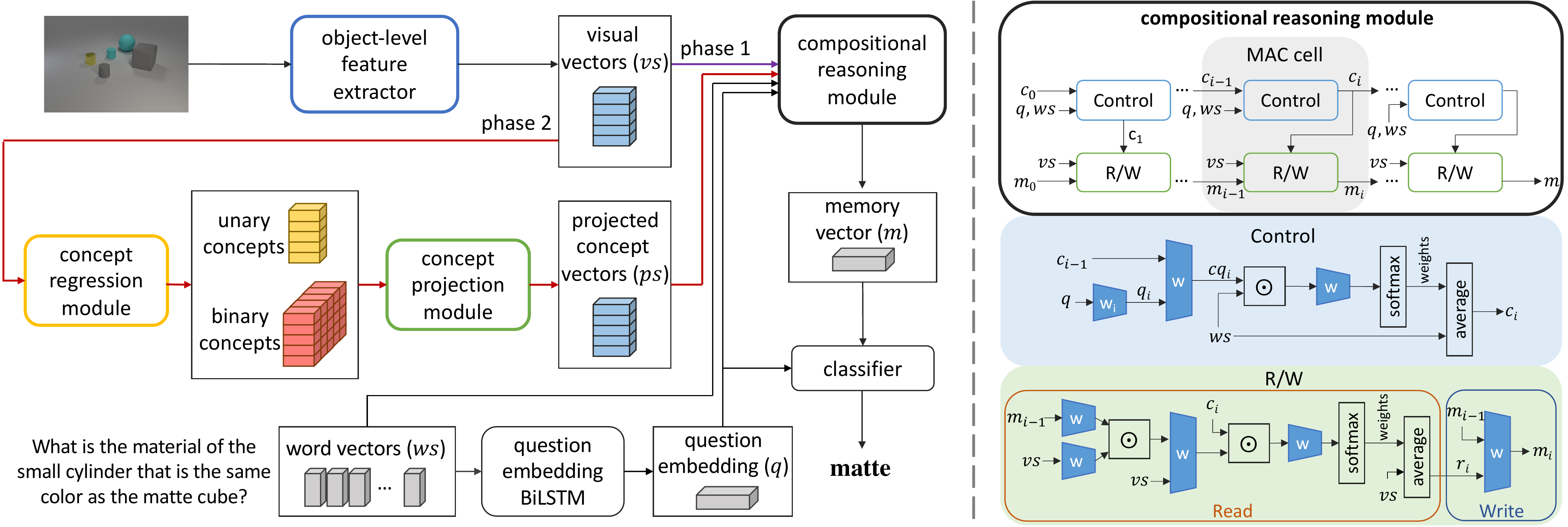}
\caption{\small{The framework and the compositional reasoning module. The left graph shows the general framework; The phase 1 training path is drawn in purple and the phase 2 training paths are drawn in red. The black paths are shared for both training phases. The structures of our proposed object-level feature extractor, concept regression module and concept projection module are shown in Figures \ref{ext}, \ref{conc_regre} and \ref{conc_proj}.}}
\label{framework}
\vspace{-3mm}
\end{figure*}

\vspace{-3mm}
\section{Related Works}
\vspace{-2mm}
\textbf{Visual Question Answering (VQA)} requires models to reason a question about an image to infer an answer. Recent VQA approaches can be partitioned into two groups: holistic models \cite{yang2016stacked, xu2016ask, anderson2018bottom, perez2018film, hudson2018compositional} and modular models \cite{andreas2016learning, andreas2016neural, hu2017learning, johnson2017inferring, hu2018explainable, yi2018neural, mao2018neuro}, according to whether the approach has explicit sub-task structures. A typical holistic model, MAC \cite{hudson2018compositional}, perform iterative reasoning steps with an attention mechanism on the image.  A modular framework, NS-CL \cite{mao2018neuro}, designs multiple principle functions over the extracted features to explain the reasoning process. 

\textbf{Scene graph grounding} requires to construct the relationship among objects in an image. \cite{yang2018graph} designs a graph R-CNN model to detect objects and classify relations among them simultaneously. \cite{bajaj2019g3raphground} uses graphs to ground words and phrases to image regions. \cite{yeh2018unsupervised} proposes to link words to image concepts in an unsupervised setting. However, all these works have predefined object and relation concepts. We focus on inducing the concepts from the language compositionality to better interpret the reasoning framework.

\textbf{Model interpretability} aims to explain the neural model predictions. \cite{bau2017network} proposed network dissection to quantify interpretability of CNNs. \cite{zhang2019interpreting} explains a CNN at the semantic level with decision trees. \cite{shi2019explainable} generates scene graphs from images to explicitly trace the reasoning-flow. \cite{park2016attentive, hu2018explainable} focused on visual attentions to provide enhanced interpretability. 
Our work is closely related to \emph{the self-explaining systems via rationalization}~\cite{lei2016rationalizing,chen2018learning,yu2019rethinking}.
These works usually extract subsets of inputs as explanations, while our work moves one-step further by learning parts of the structural explanation definitions (i.e., our concept hierarchy) together with explanations (i.e., the concept-level reasoning flow).

\textbf{Visual concept learning} contributes to broad visual-linguistic applications, such as cross-modal retrieval \cite{kiros2014unifying}, visual captioning \cite{karpathy2015deep}, and visual-question answering \cite{malinowski2015ask, antol2015vqa}. \cite{yi2018neural, mao2018neuro} attempt to disentangle visual concept learning and reasoning. Based on the visual concepts learned from VQA, \cite{han2019visual} learns metaconcepts, i.e., relational concepts about concepts, with augmented QA-pairs about metaconcepts. Our work differs from the previous ones in learning concepts and super concepts without external knowledge.
\vspace{-3mm}


\section{OCCAM: Object-Centric Visual Reasoning}
\vspace{-2mm}
This section introduces a new neural architecture, object-centric compositional attention model (OCCAM), that performs visual reasoning over the object-level visual features.
This model not only achieves state-of-the-art performance, but also plays a key role in inducing object-wise or relational concepts as will be described in section \ref{sec:concept_induction}.

Figure \ref{framework} shows our general framework with two training phases, each consists to the process of attaining the answers from the input images and questions.
Phase 1 (black-colored paths) corresponds to the training of our OCCAM, in which we train the object-level feature extractor, the compositional reasoning module and the question embedding LSTM. 
Phase 2 (\textcolor{red}{red}-colored paths) corresponds to the induction of symbolic concepts based on the aforementioned trained neural modules, as well as the training of a concept projection module so that the induced concepts can be accommodated in the OCCAM pipeline. 
The figure shows the central role that the OCCAM plays in our framework.




\vspace{-2mm}
\subsection{Background on compositional reasoning}
\vspace{-2mm}
\label{ssec:mac}
\textbf{Notations.} As shown in Figure \ref{framework}, we name the visual vectors as $vs$, the output memory vector from the compositional reasoning module as $m$, the embedded word vectors for questions as $ws$, and the question embedding as $q$. 


\textbf{The compositional reasoning framework} follows a VQA setting: given a question and an image as inputs, the model is required to return the correct answer choice. The target function can thus be written as:
\begin{equation}
\small
\vspace{-2mm}
\begin{split}
    \mathfrak{L}(ws, vs, q) = -\sum_{k \in K} y_k \log \mathcal{F}(q_k, \mathcal{G}(ws_k, vs_k, q_k)) \\
    q_k = \mathcal{Q}(ws_k), vs_k = \mathcal{I}(im_k).
    \label{loss_fn}
\end{split}
\vspace{-2mm}
\end{equation}
$K$ is the total number of image-question pairs, $y$ is the one-hot ground truth vector, $\mathcal{F}$ is the classifier, $\mathcal{G}$ is the reasoning module, $\mathcal{Q}$ is the question embedding LSTM, $\mathcal{I}$ is the visual feature extractor and $im$ is the image input.


\begin{figure}[t]
\centering
\includegraphics[width=0.45\textwidth]{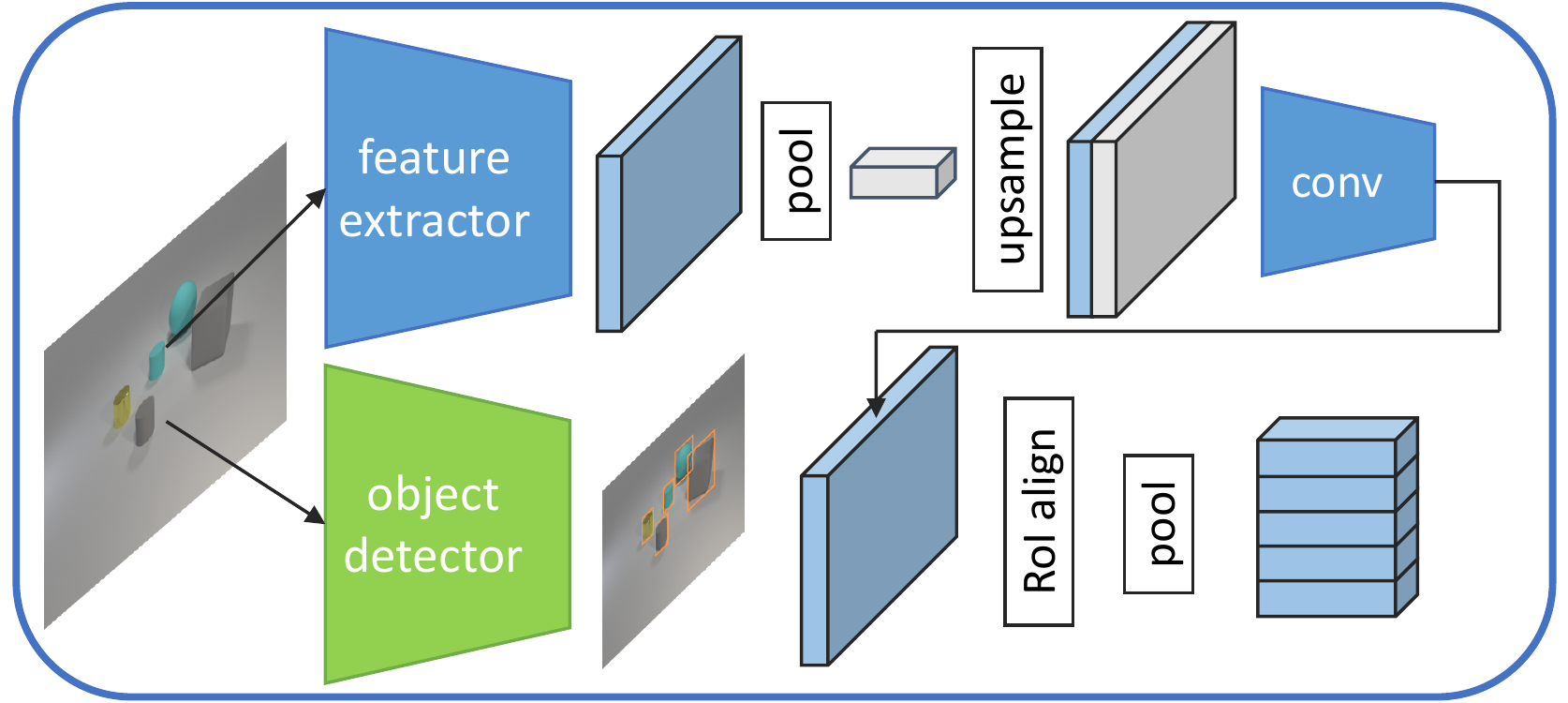}
\caption{\small{The architecture of our \textbf{object-level feature extractor}.}}
\label{ext}
\vspace{-5mm}
\end{figure}

\textbf{The MAC reasoning module} \cite{hudson2018compositional}
processes visual and language inputs in a sequential way. 
Shown in Figure \ref{framework} (right), each MAC cell contains a control unit that uses word embedding to control what object features should be read and written into memory and an R/W (Read/Write) unit that performs reading and writing object features; the blue diagrams labeled with $w$ stand for fully connected layers and the symbol $\odot$ stands for Hadamard product. 
More details can be found in Appendix \ref{app:reasoning_details}.
\vspace{-2mm}
\subsection{Object-centric compositional attention model}
\vspace{-2mm}
\label{ssec:object_mac}
Our OCCAM network is shown in Figure \ref{framework} with phase 1 path.
It performs MAC-style reasoning, but over the object-level visual features generated by our proposed \textbf{object-level feature extractor} (Figure \ref{ext}).
Fed with an image, the extractor produces a set of vectors $vs$, each encodes a single object's unary visual features and its interactions with other objects. The module works as the following steps:

(1) Following \cite{mao2018neuro}, we use Mask-RCNN \cite{he2017mask} to detect all objects in an image and output the bounding boxes for them. 
The image is fed to a ResNet34 network \cite{He_2016_CVPR} pretrained on ImageNet \cite{deng2009imagenet} to generate the feature maps.



(2) On top of the ResNet34 feature maps, we apply a global average pooling to get a single \textbf{global feature vector} (the gray vector in the figure). We concatenate this global vector with the feature map at each location, followed by {three convolution layers}.
This global vector is crucial since it allows the visual features to encode the interaction among objects; and the three convolution layers {fuse} the local and global features into a single visual vector at each position.

(3) Finally, to get object-level features from the above pixel-level fused features, we use RoI align \cite{he2017mask} to project the objects' bounding boxes onto the fused feature vectors to generate the RoI feature maps; and average pool these RoI maps for each object to produce the object-level $vs$.

Our object feature extractor is jointly optimized with the reasoning module with Eqn (\ref{loss_fn}) in the phase 1 training.

\vspace{-3mm}
\section{Concept Induction and Reasoning}
\vspace{-2mm}
\label{sec:concept_induction}

This section describes how we achieve our goal of inducing symbolic concepts for objects and performing compositional reasoning on the induced concepts. 
First, we formalize the problem of concept induction~(section \ref{ssec:concept_formulation}).
Second, built on the learned OCCAM network introduced in the previous section, we propose to induce concepts of both unary object properties or the binary relations between objects~(section \ref{ssec:concept_induction}). Finally, we propose compositional reasoning over symbolic concepts by substituting the object-level features with the induced concepts (section \ref{ssec:concept_reasoning}). 

\vspace{-2mm}
\subsection{Problem definition}
\vspace{-2mm}
\label{ssec:concept_formulation}
We consider identifying three types of concepts: (1) the \textbf{unary concepts} $\mathcal{C}^u$ that are properties of objects (e.g., \emph{red}, \emph{cube}, etc.); (2) the \textbf{binary concepts} $\mathcal{C}^b$ that are relation descriptions between any two objects (e.g., \emph{left}, \emph{front} etc.); and (3) the \textbf{super concepts} $\mathcal{C}^{sup}$ that are hypernyms of certain subsets of concepts (e.g., \emph{color}, \emph{shape}, etc.), subject to that each object can only possess one concept under each super concept, e.g., \emph{cube} and \emph{sphere}.

As questions refer to objects and describe object relations in images and, more importantly, include all the semantic information to reach an answer, it is natural to induce the concepts from question words. Therefore we assume that all the unary and binary concepts have their corresponding words; and these words are a subset of the nouns or adjectives from all the training questions. We denote the sets of words that describe unary concepts and binary concepts as $M^u$ and $M^b$ respectively. Therefore, the goal of concept induction consists of the following tasks:

\noindent$\bullet$ \textbf{Visual mapping:} for each concept $c \in \mathcal{C}^u$ or $\mathcal{C}^b$, learning a mapping from the visual feature $v$ to $c$. In other words, a prediction function $f_c(v) \in \{0,1\}$ is learned to predict the existence of concept $c$ from the visual feature $v$ of an object.

\noindent$\bullet$ \textbf{Word mapping:} for each concept $c \in \mathcal{C}^u$ or $\mathcal{C}^b$, identifying a subset of words $S_c \subset M^u$ or $M^b$ that are synonyms representing the same concept, e.g., the concept of \emph{'cube'} corresponds to set of words $\{cube, cubes, block, blocks, ...\}$.

\noindent$\bullet$ \textbf{Super concept induction:} clustering of concepts to form super concepts. Each super concept $\mathbf{c}$ contains a set of concepts $\{c_1, \cdots, c_k\}$ $\subset \mathcal{C}^u$ or $\mathcal{C}^b$.



\begin{algorithm}[t!]
\SetAlgoLined
\small
 \caption{\small{Classifier training data generation. ST($\cdot$) splits a vector $\alpha$$\in$$\mathbb{R}^\beta$ to a set of $\beta$ values. GMM($\cdot$) uses Gaussian Mixture Model to cluster a set of data points. FB($\cdot$) finds the decision boundary for the 2 Gaussian components. $\mathbbm{1}$ is the indicator function.}}
 \label{dp_gen}
\KwResult{$P^u$, $P^b$}
 $P^u=\{\}$, $P^b=\{\}$\\
 \For{$x \in M^u \cup M^b$}{$S_x=\{\}$, $bd_x=0$} 
 \For{$vs,ws \in$ DATASET}{
    \For{$c_w \in ws \cap M^u$}{
        $S_{c_w} = S_{c_w} \cup$ ST$(\mathcal{R}(vs,c_{w},m_0))$
    }
    \For{$c_w \in ws \cap M^b$}{
        \For{$v \in vs$}{
            $S_{c_w} = S_{c_w} \cup$ ST$(\mathcal{R}(vs,c_{w},\mathcal{W}(m_0,v)))$
        }
    }
 }
 \For{$x \in M^u \cup M^b$}{$bd_x=$ FB$($GMM$(S_x))$}
 \For{$vs,ws \in$ DATASET}{
     \For{$v_1 \in vs$}{
        \For{$c_w \in ws \cap M^u$}{
            $y = \mathbbm{1}(\mathcal{R}(v_1,c_{w},m_0) > bd_{c_w})$\\
            $P^u = P^u \cup \{(v_1, c_w, y)\}$
        }
        \For{$c_w \in ws \cap M^b$}{
            \For{$v_2 \in \{vs-v_1\}$}{
                $y = \mathbbm{1}(\mathcal{R}(v_1,c_{w},\mathcal{W}(m_0,v_2)) > bd_{c_w})$\\
                $P^b = P^b \cup \{(v_1, v_2, c_w, y)\}$
            }
        }
    }
 }
\end{algorithm}


\begin{figure}[t!]
\centering
\includegraphics[width=0.45\textwidth]{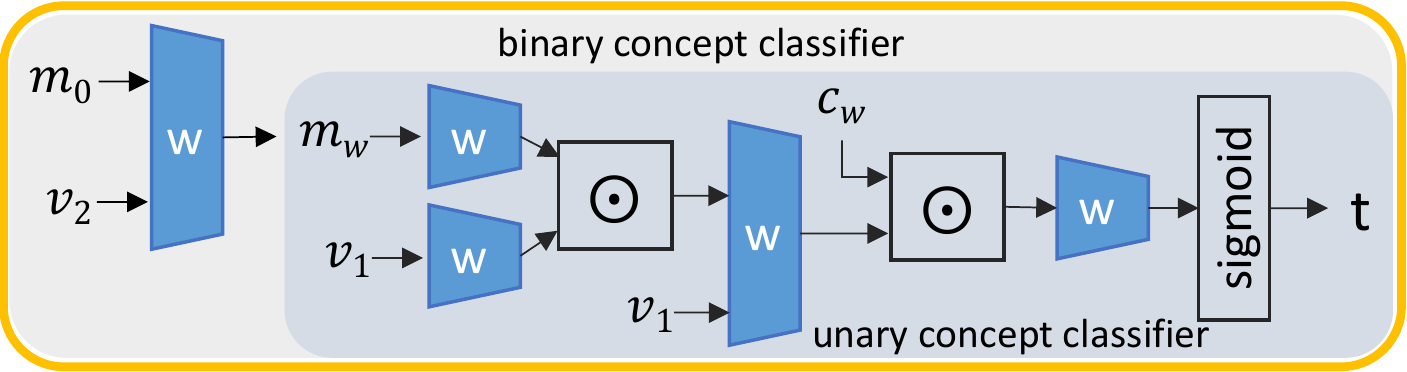}
\caption{\small{The structure of the \textbf{concept regression module}. $v_1$ and $v_2$ are the object-level visual vectors representing two objects respectively, and $c_w$ is the word vector. $m_0$ is a fixed vector and $m_w$ equals to $m_0$ for the unary concept classifier.}}
\label{conc_regre}
\vspace{-5mm}
\end{figure}

\begin{figure*}
\centering
\includegraphics[width=0.99\textwidth]{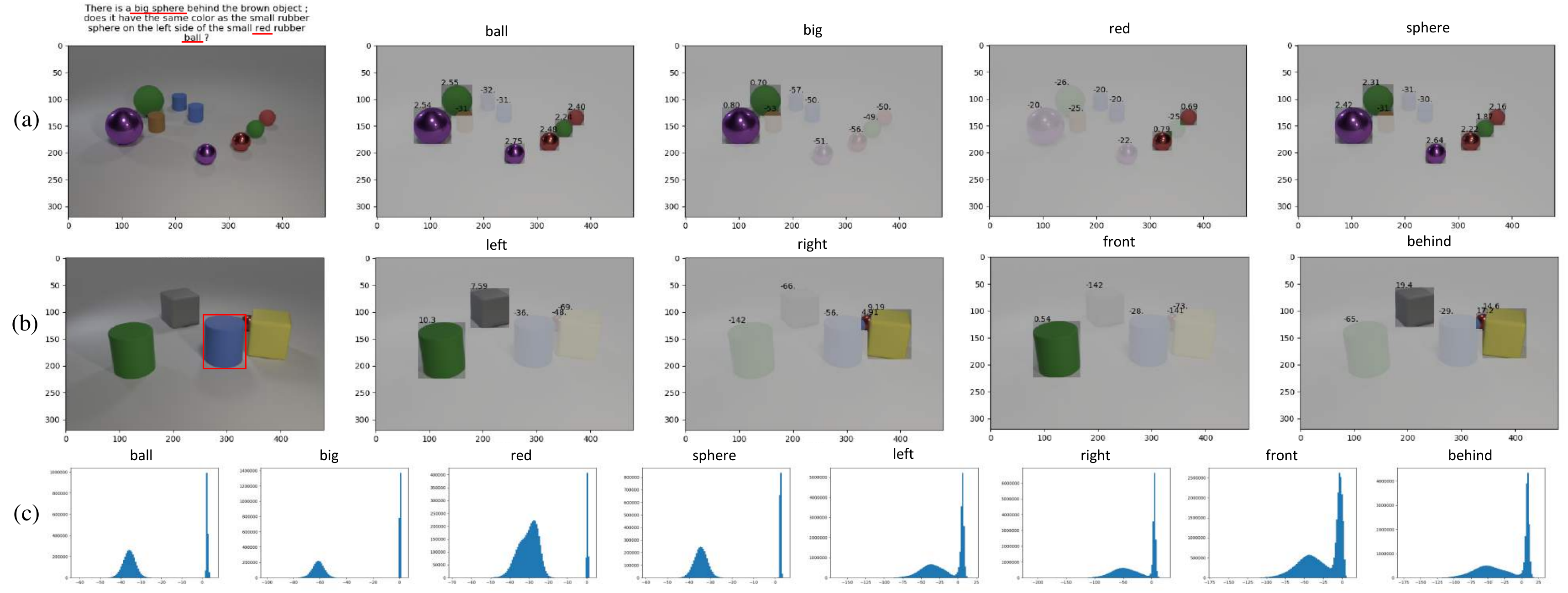}
\caption{\small{Attention visualization and attention logit distributions. (a) The attention visualization corresponding to the words describing the unary concepts by performing $\mathcal{R}(vs,c_{w},m_0)$. Each of the words above the latter 4 images corresponds to a unique $c_w$ and the value on each object is the attention logit (the same applies to (b)). (b) The attention visualization corresponding to the words describing the binary concepts by performing $\mathcal{R}(vs,c_{w},\mathcal{W}(m_0,v_2))$. $v_2$ represents the object bounded by a red rectangle in the first image. (c) the attention logit distribution corresponding to each word describing a concept. }}
\label{attn_vis}
\vspace{-6mm}
\end{figure*}

\subsection{Concept induction}
\vspace{-2mm}
\label{ssec:concept_induction}
This section describes how we achieve the aforementioned tasks of concept induction. The key idea of our approach includes: (1) benefiting from the R/W unit from the trained MAC cells to achieve the visual mapping to textual words; (2) utilizing the inclusiveness of words' visual mapping to induce each concept's multiple word descriptions; (3) clustering super concepts from the mutual exclusiveness between concepts.
To achieve the above, we first train two binary classifiers that can determine if a word correctly describes an object's unique feature and if a word correctly describes a relation between two objects respectively. Then, with the help of these classifiers, we produce zero-one vectors for words that properly describe the unique features for each object and the relations between any pair of objects in single images across the dataset. Finally, we perform a clustering method on the word vectors to generalize unary and binary concepts, and the super concept sets.

\vspace{-5mm}
\paragraph{Visual mapping via regression from MAC cells}
The concept regression module is shown in Figure \ref{conc_regre}. It is composed of a classifier for the unary concept word regression, $\mathcal{B}^u(v_1, c_w) \in [0,1]$, and a classifier for the binary concept word regression, $\mathcal{B}^b(v_1, v_2, c_w) \in [0,1]$. $\mathcal{B}^u$ is expected to produce 1 if $v_1$ can be described by the word vector $c_w$. Likewise, $\mathcal{B}^b$ is expected to produce 1 if the relation of $v_1$ to $v_2$ can be described by the word vector $c_w$ . 

We generate training data points $P^u=\{({v_1}_i^u,{c_w}_i^u,y_i^u)\}$ and $P^b=\{({v_1}_i^b,{v_2}_i^b,{c_w}_i^b,y_i^b)\}$ for $\mathcal{B}^u$ and $\mathcal{B}^b$ by utilizing the Read/Write unit (Figure \ref{framework}(right)) in the reasoning module after phase 1 training.
The whole generation process is described in Algorithm \ref{dp_gen}.
We denote $\mathcal{R}(vs,c_{i},m_{i-1}) \in \mathbb{R}^{|O|}$ for the sequence of functions before the softmax operation in the Read unit and $\mathcal{W}(m_{i-1},r_i) \in \mathbb{R}^D$ for the function of the Write unit, where $O$ is the set of objects in an image and $D$ is the vector dimension.

Specifically, our algorithm first uses $\mathcal{R}(\cdot,\cdot,\cdot)$ and $\mathcal{W}(\cdot,\cdot)$ to find the attention logits on the objects corresponding to words describing the unary and binary concepts in a question as shown in Figure \ref{attn_vis}(a\&b). 
We then use the values of logits to determine if the object possesses the concept of the word (positive) or not (negative).
Noticing the attention logit distribution of the sampled objects for each word is a two-peak distribution (Figure \ref{attn_vis}(c)), we use a GMM \cite{xuan2001algorithms} with two Gaussian components to model the distribution and find the decision boundary for each word's attention logit distribution. Observe that the distribution of a binary concept word has two interfering waves, because in some cases it is hard to tell if two objects have that relation ('front' is inappropriate if two objects are on the same horizon).
$P^u$ and $P^b$ are generated by classifying the data points to positives and negatives with the decision boundaries. 
Finally, we can train $\mathcal{B}^u$ and $\mathcal{B}^b$ with data $P^u$ and $P^b$ by minimizing the binary cross entropy loss.


\vspace{-5mm}
\paragraph{Binary coding of objects}
With trained $\mathcal{B}^u$ and $\mathcal{B}^b$, we represent an object $o_1$ with a binary code vector. Each dimension corresponds to a word. A dimension has value 1 if the corresponding word can describe $o_1$ and 0 otherwise. The binary vectors of object properties and of the relations between two objects, $o_1$ and $o_2$ can be computed with the functions $\gamma^u \in \mathbb{R}^{|M^u|}$ and $\gamma^b \in \mathbb{R}^{|M^b|}$ respectively:
\vspace{-2mm}
\begin{equation}
\small
\begin{split}
    \gamma^u = \mathbbm{1}_{i>0.5}(\mathcal{B}^u(v_1,C^u)) \\
    \gamma^b = \mathbbm{1}_{i>0.5}(\mathcal{B}^b(v_1,v_2,C^b)),
    \label{word_represent}
\end{split}
\vspace{-2mm}
\end{equation}
where $v_1$ and $v_2$ are the object-level visual vectors of $o_1$ and $o_2$, $C^u \in \mathbb{R}^{|M^u| \times D}$ and $C^b \in \mathbb{R}^{|M^b| \times D}$ are the stacks of word embeddings in vocabulary $M^u$ and $M^b$. $\mathbbm{1}_\alpha(\beta)$ performs elementwise on $\beta$: return 1 if the element satisfies condition $\alpha$ or 0 otherwise. 

\begin{figure}
\centering
\includegraphics[width=0.45\textwidth]{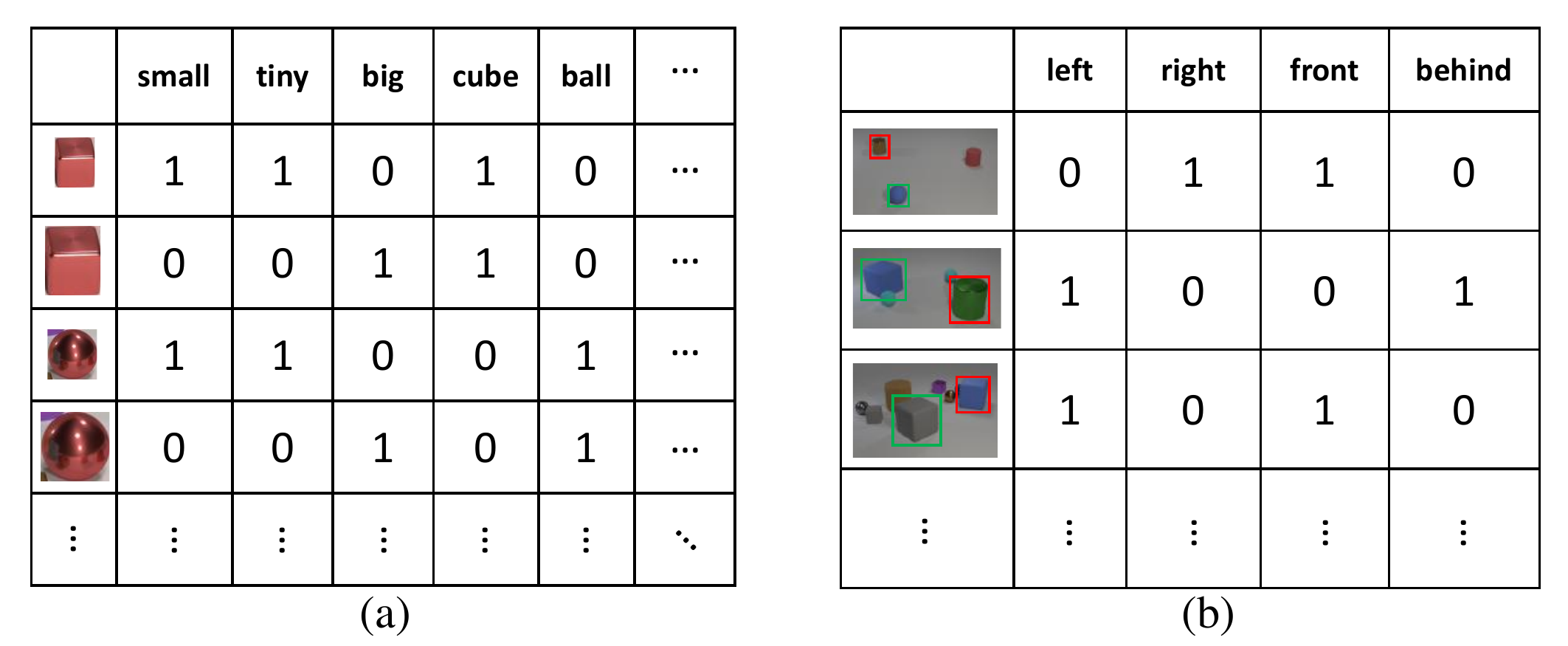}
\caption{\small{The zero-one matrices indicating word descriptions of objects and object relations across the dataset. (a) The matrix $\Gamma^u$ indicates what words can describe objects. (b) The matrix $\Gamma^b$ indicates what words can describe the relations object $v_1$'s (bounded by green rectangles) are to object $v_2$'s (bounded by red rectangles).}}
\label{cls_vector}
\vspace{-8mm}
\end{figure}

By applying $\gamma^u$ and $\gamma^b$ to all the objects and relations in the dataset, we can attain a matrix $\Gamma^u \in \{0,1\}^{M^u,N^u}$ and a matrix $\Gamma^b \in \{0,1\}^{M^b,N^b}$ as shown in Figure \ref{cls_vector}, where $N^u$ and $N^b$ are the total numbers of objects and co-occurred object pairs. The two matrices summarize each word's corresponding visual objects in the whole dataset.

\vspace{-6mm}
\paragraph{Concept/super-concept induction}
Finally, we group synonym words to unary and binary concepts and generate the super concepts. 
These two tasks are achieved via exploring the word inclusiveness and the concept exclusiveness captured by $\Gamma^u$ and $\Gamma^b$:
(1) words describing the same concept correspond to similar column vectors, e.g., $\Gamma^u_{small}$ and $\Gamma^u_{tiny}$;
(2) words describing exclusive concepts have column vectors that usually do not have 1 values on same objects simultaneously, e.g., $\Gamma^u_{cube}$ and $\Gamma^u_{ball}$. 
Based on the aforementioned ideas, we define the correlation metric between two words $c_{w_1}$ and $c_{w_2}$ as below:
\vspace{-4mm}
\begin{multline}
\small
    \theta_{c_{w_1}, c_{w_2}} = P(\gamma_{c_{w_1}}=1 \mid \gamma_{c_{w_2}}=1) + \\
    P(\gamma_{c_{w_2}}=1 \mid \gamma_{c_{w_1}}=1) \\
    = \frac{|\Gamma_{c_{w_1}} \odot \Gamma_{c_{w_2}}|_1^1}{|\Gamma_{c_{w_2}}|_1^1}+\frac{|\Gamma_{c_{w_1}} \odot \Gamma_{c_{w_2}}|_1^1}{|\Gamma_{c_{w_1}}|_1^1}.
    \label{corr}
\vspace{-2mm}
\end{multline}
This guarantees that $\theta \rightarrow 0^+$ for two synonym words, $\theta \rightarrow 2^-$ for two words corresponding to exclusive concepts and $\theta \in (0,2)$ for words corresponding to different nonexclusive concepts. We can produce the correlation sets for the words describing the unary concepts and the binary concepts respectively with Eqn (\ref{corr_set}).
\begin{equation}
\vspace{-2mm}
\small
    \Theta^x = \{\theta_{c_{w_1}, c_{w_2}}\}; c_{w_1}, c_{w_2} \in M^x; x \in\{u,b\} \\
    \label{corr_set}
\end{equation}

Our final step fits two GMM on $\Theta^u$ and $\Theta^b$ respectively. Each GMM has three components $\mathcal{N}_0$, $\mathcal{N}_1$ and $\mathcal{N}_2$, with their mean values initialized with 0,1 and 2. 
We then induce the unary and binary concepts, where each concept consists of synonym words whose mutual correlation is clustered to the Gaussian component $\mathcal{N}_0$. 
Similarly, we induce the super concepts, where each super concept contains multiple concepts and any two words from different concepts have correlation clustered to the Gaussian component of $\mathcal{N}_2$.

We denote the set of words corresponding to a concept $e$ as $\rho_e$, the set of the super concept sets as $L$, the set of all concepts as $E$. Then, we can represent all the objects in an image with a unary concept matrix $K^u$ and represent all the relations between any two objects in an image with a binary concept matrix $K^b$ with Algorithm \ref{conc_gen}.

\begin{algorithm}[t]
\small
\SetAlgoLined
\KwResult{$K^u$, $K^b$}
 $K^u=\textbf{0}^{|O| \times |E^u|}$, $K^b=\textbf{0}^{|O| \times |O| \times |E^b|}$\\
 \For{$i \in O$}{
    \For{$e^u \in E^u$}{
        $K^u[i][e^u] = $MAX$(\mathcal{B}^u(v_i, \rho_{e^u})$)\\
    }
    \For{$l^u \in L^u$}{
        $K^u[i][l^u] = $HARDMAX$(K^u[i][l^u])$\\
    }
    \For{$j \in O-\{i\}$}{
        \For{$e^b \in E^b$}{
            $K^b[i][j][e^b] = $MAX$(\mathcal{B}^b(v_i, v_j, \rho_{e^b}))$\\
        }
        \For{$l^b \in L^b$}{
            $K^b[i][j][l^b] = $HARDMAX$(K^u[i][j][l^b])$\\
        }
    }
 }
 \caption{\small{Concept vector generalization. MAX$(\alpha)$ and HARDMAX$(\alpha)$ return the largest value in vector $\alpha$ and its position as a one-hot vector, respectively.}}
 \label{conc_gen}
\end{algorithm}

\vspace{-1mm}
\subsection{Concept compositional reasoning}
\vspace{-1mm}
\label{ssec:concept_reasoning}
\begin{figure}
\centering
\includegraphics[width=0.45\textwidth]{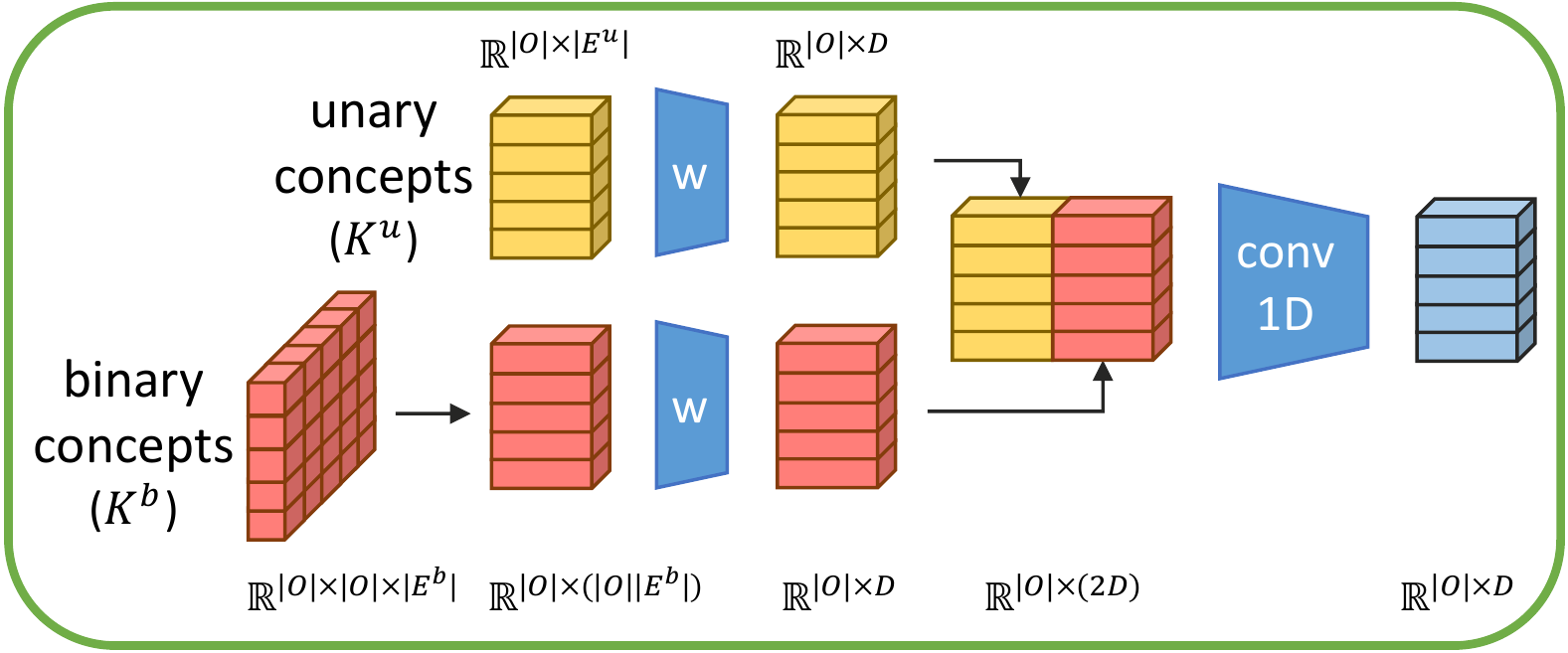}
\caption{The structure of the \textbf{concept projection module}. We label the dimensions of matrices near them in the graph.}
\label{conc_proj}
\vspace{-6mm}
\end{figure}

Our ultimate goal is to perform compositional reasoning to answer a question with the generated concept representations $K^u$ and $K^b$ for an image; so as to confirm that our induced concepts are accurate and sufficient. We achieve this with the phase 2 training process in Figure \ref{framework}.
The key idea is to transplant the learned compositional reasoning module from manipulating the visual features to manipulating $K^u$ and $K^b$, for attaining the answer to a question.

\begin{figure*}
\captionof{table}{
\small{The comparison of our OCCAM framework to the state-of-the-art methods on CLEVR (left) and GQA (right) datasets. $\dagger$ means training with additional program supervision. $\ddagger$ means pretraining on larger visual+language corpora. $\S$ means pretraining a scene-graph extraction model with additional rich annotated data.}}
\vspace{-2mm}
\label{sota_comp}

  \begin{minipage}[b]{0.55\textwidth}
\small
\centering
(a) CLEVR
\begin{tabular}{l|c|p{0.45cm}p{0.45cm}p{0.53cm}p{0.53cm}p{0.53cm}}
        \toprule
         method & overall & count & exist & comp num & query attr & comp attr  \\
         \midrule
         RN \cite{santoro2017simple} & 95.5 & 90.1 & 93.6 & 97.8 & 97.1 & 97.9 \\
         FiLM \cite{perez2018film} & 97.6 & 94.5 & 93.8 & 99.2 & 99.2 & 99.0 \\
         MAC \cite{hudson2018compositional} & 98.9 & 97.2 & 99.4 & \textbf{99.5} & 99.3 & 99.5 \\
         NS-CL \cite{mao2018neuro} & 98.9 & \textbf{98.2} & 99.0 & 98.8 & 99.3 & 99.1\\
         OCCAM (ours) & \textbf{99.4}&98.1&\textbf{99.8}&99.0&\textbf{99.9}&\textbf{99.9}\\
         \midrule
         NS-VQA$^{\dagger}$ \cite{yi2018neural} & {99.8} & {99.7} & {99.9} & {99.9} & {99.8} & {99.8}\\ 
         \midrule
         Human \cite{johnson2017inferring} & 92.6 & 86.7 & 96.6 & 86.5 & 95.0 & 96.0 \\
         \bottomrule
    \end{tabular}
  \end{minipage}
  \begin{minipage}[b]{0.37\textwidth}
  \small
    \centering
    (b) GQA
    \small
    \begin{tabular}{l|c|c|c}
        \toprule
         method  & val & test-dev & test  \\
         \midrule
         MAC \cite{andreas2016neural} & 57.5 & - & 54.1 \\
         LXMERT \cite{tan2019lxmert}  & - & 50.0 & - \\
         LCGN \cite{hu2019language}  & 63.9 & 55.8 & 56.1 \\ 
         OCCAM (ours)  & \bf 64.5 & \bf 56.2 & \bf 56.2\\ 
         \midrule
         MMN \cite{chen2021meta}$^{\dagger}$ & - & 60.4 & 60.8 \\
         NSM \cite{hudson2019learning}$^{\S}$ & - & 63.0 & 63.2 \\
         LXMERT \cite{tan2019lxmert}$^{\ddagger}$ & - & 60.0 & 60.3 \\
         ViLT \cite{kim2021vilt}$^{\ddagger\S}$  & - & 65.1 & 64.7 \\
         \bottomrule
    \end{tabular}
    \end{minipage}
    \vspace{-3mm}
  \end{figure*}

To this end, first, we project $K^u$ and $K^b$ to the same vector space with $vs$ with the concept projection module shown in Figure \ref{conc_proj}, so that the compositional module can perform the reasoning steps on the projected concept vectors.
Specifically, we first reduce the dimension of $K^b$ from $\mathbbm{R}^{|O| \times |O| \times |E^b|}$ to $\mathbbm{R}^{|O| \times |O||E^b|}$, resulted in $\hat{K}^b$, because $K^b$ can be understood as the relations to other objects for each object in an image. Then, we use two separate fully connected networks to project $K^u$ and $\hat{K}^b$ respectively, concatenate and use a sequence of 1D convolution layers to project the results to the same dimension of $vs$'s. 

Second, to minimize the discrepancy between the distribution of our projected vectors and that of the original visual vectors $vs$, we fix the weights of other modules in the framework and only train the concept project module by optimizing the target function Eqn. (\ref{loss_fn}). Then, we train the concept projection module and the compositional reasoning module with other modules' weights fixed to better optimize Eqn. (\ref{loss_fn}). The result is a compositional reasoning model that works on the induced concepts only.

\vspace{-3mm}
\section{Experiment}
\vspace{-2mm}
\subsection{Settings}
\vspace{-2mm}
\paragraph{Datasets}
(1) We first evaluate our model on the \textbf{CLEVR} \cite{johnson2017clevr} dataset. The dataset comprises images of synthetic objects of various shapes, colors, sizes and materials and question/answer pairs about these images. The questions require multi-hop reasoning, such as finding the transitive relations, counting numbers, comparing properties, to attain correct answers. 
Each question corresponds to a ground truth human-written programs.
Because the programs rely on pre-defined concepts thus do not fit our problem, we let our framework learn from scratch \emph{without} using the program annotations.
There are 70k/15k images and $\sim$700k/$\sim$150k questions in the training/validation sets. We follow the previous works \cite{yi2018neural,hudson2018compositional,mao2018neuro} to train our model on the whole training set and test on the validation set. 

(2) To demonstrate the generalizability of our approach, we further evaluate on the \textbf{GQA} dataset.
GQA is a real-world visual reasoning benchmark. It consists of 113K images collected from the Visual Genome dataset \cite{krishna2017visual} and 22M questions. It has a train split for model training and three test splits (val, test, test-dev) \cite{hudson2019gqa}. The dataset provides the detected object features extracted from a Faster RCNN detector \cite{ren2015faster}, so each object is represented as a 2048-dimensional vector.

\vspace{-4mm}
\paragraph{Implementation details} We include the checklist of our implementation details in Appendix~\ref{app:training_details}.

\begin{table}[t!]
\small
    \centering
    \caption{\small{Effect of the choice of reasoning steps for our model.}}
    \begin{tabular}{l|cccc}
        \toprule
         steps & 4 & 8 & 12 & 16  \\
         \midrule
         accuracy (CLEVR) & 94.3 & 98.6 & \textbf{99.4} & 99.1 \\
         accuracy (GQA test-dev) & 55.1 & 55.6 & 55.2 & \textbf{56.2} \\
         \bottomrule
    \end{tabular}
    \label{ablation}
    \vspace{-3mm}
\end{table}
\subsection{Object-level reasoning}
We first perform the end-to-end phase 1 training shown in Figure \ref{framework}, i.e., our OCCAM model. The performance comparison of our model to the state-of-the-art models is shown in Table \ref{sota_comp}. Under the setting that no external human-labeled programs and no pretraining are used, our model achieves state-of-the-arts compared with published results on both CLEVR and GQA datasets. For comparison with models on GQA leaderboard, we also train our OCCAM model on train-all split and achieves an accuracy of 58.5\% on the test-standard split of GQA dataset, which outperforms other popular models (e.g. MCAN, BAN and LCGN) trained with no additional data (accuracies are 57\%$\sim$58\%). While transformer-based methods with pretraining phase boost the performance, however, they undermine the model’s explainability and make it difficult to induce concepts. On CLEVR, our model also has an on-par performance with the best model \cite{yi2018neural} that uses external human-labeled programs.

Compared to the original MAC \cite{hudson2018compositional} framework which uses image-level attentions, our model proves that the constraint of attentions on the objects are useful for improving the performance on both datasets, with significant improvement on the validation sets. We do not use the position embedding to explicitly encode the positions of objects for relational reasoning; however, we use the global features to enhance the model's understanding of inter-object relations. This shows that the relations among objects are learnable concepts without external knowledge for the deep network. 

Table \ref{ablation} further gives an ablation study on the numbers of reasoning steps, i.e., the number of MAC modules, for our model. The reasoning model with 4 steps has a performance gap to the models with 8, 12 or 16 steps, while the latter three models have on-par performances. We conjecture that the model with low reasoning steps may not be able to capture multiple hops of a question and the model performance converges with an increasing number of reasoning steps. We also did ablation study on the contribution of object-level feature extractor on CLEVR dataset. With pretrained ResNet101 features, learnable ResNet34 features, learnable ResNet34 features plus global features respectively, the model achieves 97.9\%, 99.0\% and 99.4\% on the validation set. It shows the importance of enhancing global context understanding at object level.

\begin{figure*}
\captionof{table}{
\small{Comparison of our visual feature-based OCCAM and our concept-only OCCAM. The number of reasoning steps is 8.}}
\vspace{-2mm}
\label{obj_conc_comp}
\begin{minipage}[b]{0.67\textwidth}
\small
\centering
(a) CLEVR
\begin{tabular}{l|c|ccccc}
        \toprule
         method & overall & count & exist & comp num & query attr & comp attr  \\
         \midrule
         OCCAM$_{\textrm{visual}}$& 98.6 & 95.9 & 99.8 & 96.2 & 99.8 & 99.7 \\
         
         OCCAM$_{\textrm{concept}}$ & 97.9&95.6&98.7&97.3&98.4&99.3\\
         \bottomrule
    \end{tabular}
  \end{minipage}
  \begin{minipage}[b]{0.3\textwidth}
  \small
    \centering
    (b) GQA
    \begin{tabular}{l|c|c}
        \toprule
         method & val & test-dev  \\
         \midrule
         OCCAM$_{\textrm{visual}}$ & 63.8 & 55.6 \\
         OCCAM$_{\textrm{concept}}$ & 63.1 & 54.2\\
         \bottomrule
    \end{tabular}
    \end{minipage}
    \vspace{-3mm}
  \end{figure*}

         
\subsection{Concept induction and reasoning}
Next we evaluate the performance of our concept induction method, i.e., the phase 2 training in Figure ~\ref{framework}.
To qualitatively show that our induced concepts capture sufficient and accurate information for visual reasoning, we replace the visual inputs to the objects' induced concepts according to Section~\ref{ssec:concept_reasoning}. The resulted model, denoted as OCCAM$_{\textrm{concept}}$, is expected to perform closely to the original OCCAM with high-quality induced concepts.

Table \ref{obj_conc_comp} gives the results.
To achieve the balance of the performance and the interpretability, we make the OCCAM model run 8 reasoning steps for both concept induction and reasoning. It is observed that our concept-based OCCAM (with induced concept features only) achieves on-par performance with the original OCCAM model (with full input visual features).
We also visualize the reasoning steps for the OCCAM$_{\textrm{concept}}$ model in Appendix \ref{app:vis_concept_reasoning}.

\begin{figure}[t]
\centering
\includegraphics[width=0.49\textwidth]{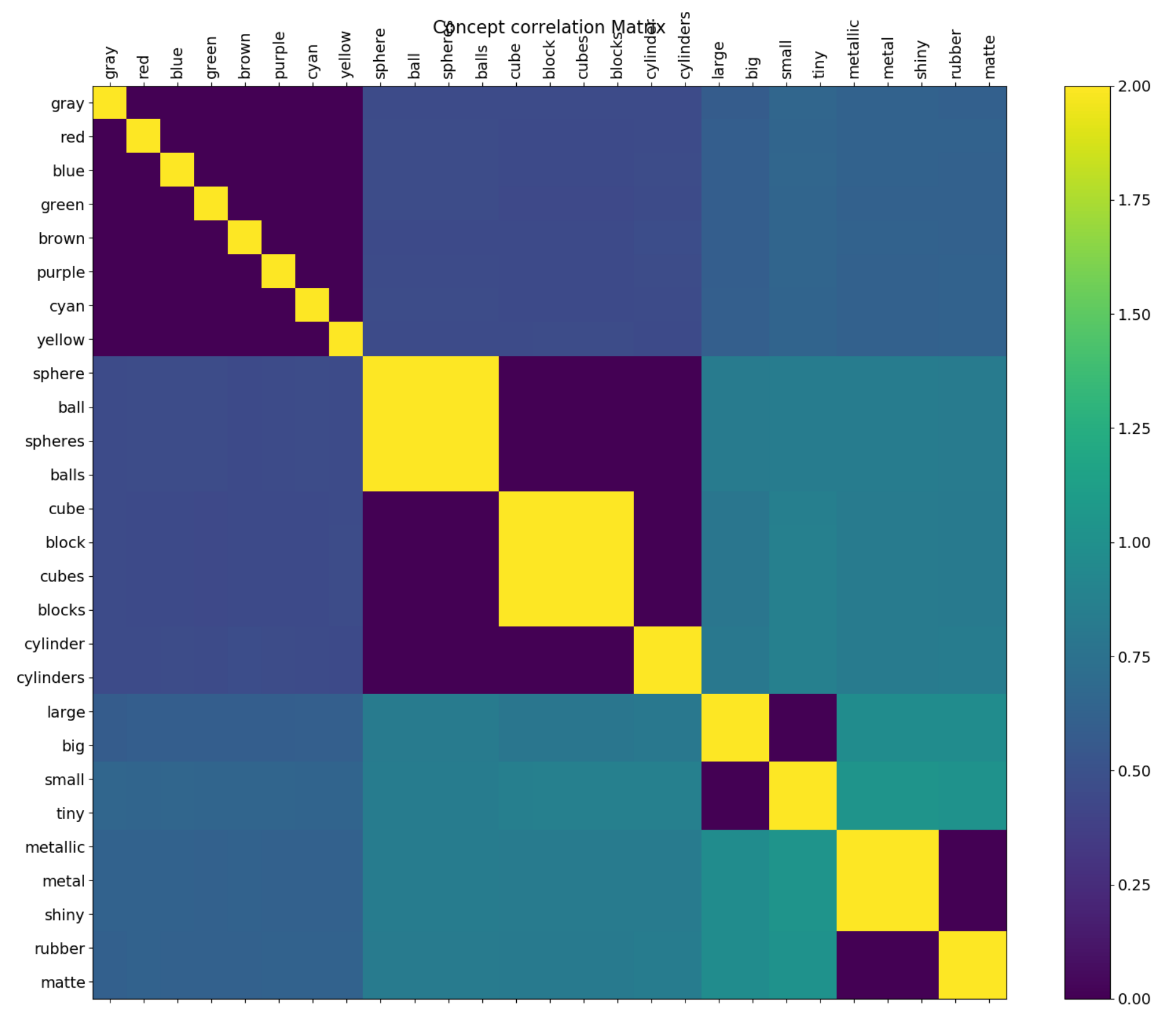}
\vspace{-3mm}
\caption{The CLEVR unary concept correlations $\Theta^u$.}
\label{conc_corr}
\vspace{-3mm}
\end{figure}
\begin{figure}[t]
\centering
\includegraphics[width=0.49\textwidth]{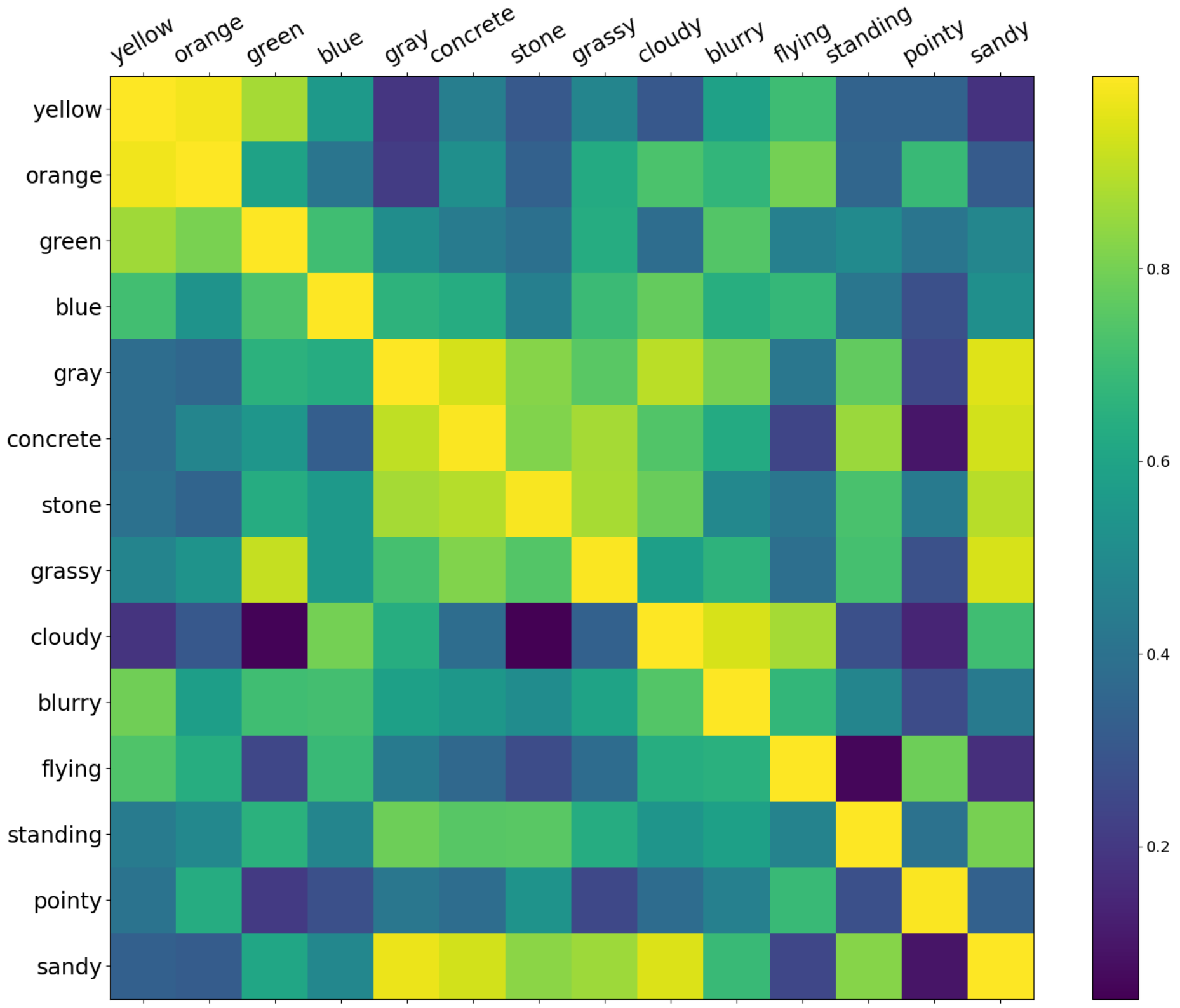}
\vspace{-3mm}
\caption{The subset of GQA concept correlations.}
\label{conc_corr_gqa}
\vspace{-5mm}
\end{figure}

\vspace{-3mm}
\paragraph{Human study of concept induction}
We present the unary concept correlations $\Theta^u$ in Figure \ref{conc_corr} and \ref{conc_corr_gqa} for both CLEVR and GQA. Since GQA consists of a huge vocabulary with highly-correlated concepts, we demonstrated a sub-set of concepts associated to general words/phrases.

On CLEVR, the concept definition from the data generator can be perfectly recovered by our approach: from Figure~\ref{conc_corr},
the correlation between any pair of synonyms is close to 2, the correlation between words belonging to the same super concept set is close to 0, and the correlation between words belonging to two different super concept sets is in the middle of the range$[0,2]$. 
%
Appendix~\ref{app:concept_correlation} provides the full generated concept hierarchy, which perfectly matches the definition in CLEVR generator and human prior knowledge, i.e., 100\% accuracy according to our human investigation.

On GQA, the correlations between words are complicated. We present a subset of word correlations in Figure \ref{conc_corr_gqa}. Instead of using Eqn. (\ref{corr}), here we show the conditional probability that a column attribute exists given that the row attribute exists. We observe that words describing a similar property have a high positive correlation, such as `yellow' and `orange', `concrete' and `stone'. They can be grouped into a single concept. Words of exclusive meanings negatively correlates with each other, such as `flying' and `standing', `pointy' and `sandy'. They can be grouped into a super concept. However, the real-world data makes it difficult to induce some commonsense super concepts. For example, the same object can have multiple colors (e.g., sky can be both gray and blue). Also, object concept can have degrees (light or dark blue), so we have to use soft values to represent the concepts. We additionally conduct human studies on the pairwise accuracy of detected concept and super concept clusters, which can be found in Appendix \ref{app:human_study}.

We provide more visualization results in Appendix \ref{app:concept_analogy} and  \ref{app:concept_dis}, including the extension of word analogy~\cite{mikolov2013distributed} (e.g., ``Madrid'' - ``Spain'' + ``France'' $\rightarrow$ ``Paris'') to multi-modality and the quantification of distance between two visual objects with the super concept space.
%


\vspace{-3mm}

\section{Conclusions}
\vspace{-2mm}
Our proposed OCCAM framework performs pure object-level reasoning and achieves a new state-of-the-art without human-annotated functional programs on the CLEVR dataset.
Our framework makes the object-word cooccurrence information available, which enables induction of the concepts and super concepts based on the inclusiveness and the mutual exclusiveness of words' visual mappings.
When working on concepts instead of visual features, OCCAM achieves comparable performance, proving the accuracy and sufficiency of the induced concepts. 
For future works, our method can be extended to more sophisticated induction tasks, such as inducing concepts from phrases, with more complicated hierarchy, with degrees of features (e.g., \emph{dark blue}, \emph{light blue}) and inducing complicated relations between objects (e.g. \emph{a little bigger}).


\paragraph{Acknowledgments}
This work is in part supported by IBM-Illinois Center for Cognitive Computing Systems Research (C3SR) - a research collaboration as part of the IBM AI Horizons Network.

{\small
\bibliographystyle{ieee_fullname}
\bibliography{egbib}
}

\clearpage
\appendix
\section{Details of compositional reasoning frameworks}
\label{app:reasoning_details}
\paragraph{Baseline visual reasoning framework} The original compositional reasoning framework \cite{hudson2018compositional} is similar to the phase 1 of our framework in Figure 2 of the main paper, except that it works on pixel-level instead of object-level features. To generate $vs$, it feeds the image to a ResNet101 \cite{He_2016_CVPR} pretrained on ImageNet \cite{deng2009imagenet} and flatten the last feature maps across the width and height as $vs$. For the question inputs, we first convert each question word to its word embedding vector ($ws$), then input $ws$ to a bidirectional LSTM \cite{hochreiter1997long,graves2005framewise} to extract the question embedding vector $q$. The compositional reasoning module takes $vs$, $ws$ and $q$ as inputs and performs multi-step reasoning to attain $m$, the final step memory output. Finally, the classifier outputs the probability for each answer choice with a linear classifier over the concatenation of $m$ and $q$. 
\vspace{-3mm}


\paragraph{The MAC reasoning module}

At each step, the $i$-th MAC cell receives the control signal $c_{i-1}$ and the memory output from the previous step, $m_{i-1}$, and outputs the new memory vector $m_i$.
The control unit computes the single $c_i$ to control reading of $vs$ in the R/W unit. Specifically, it computes the interactions among $c_{i-1}$, $q_i$, and each vector in $ws$ to produce the attention weights, and weighted averages $ws$ to produce $c_i$. The control unit of each MAC cell has a unique question embedding projection layer, while all other layers are shared.
The R/W unit aims to read the useful $vs$ and store the read information into $m_i$. It first computes the interactions among $m_{i-1}$, $c_{i-1}$ and each vector in $vs$ to attain the attention weights, weighted averages $vs$ to produce a read vector $r_i$, and finally computes the interaction of $r_i$ and $m_{i-1}$ to produce $m_i$.  The weights of the R/W units are shared across all MAC cells. 
The initial control signal and memory $c_0$ and $m_0$ are learnable parameters. 


\section{Implementation details}
\label{app:training_details}

\paragraph{CLEVR}
We set the hidden dimension $D$ to 512 in all modules. We follow \cite{hudson2018compositional} to design the question embedding module, the compositional module and the classifier. For the object-level feature extracter, we make the backbone ResNet34 learnable and zero-pad the output $vs$ to 12 vectors in total for any image. Notice that the maximum number of objects in an image is 11, so that the reasoning module is able to read nothing into the memory for some steps. For the concept projection module, to cover the full view of $vs$, the conv1D consists of five 1D convolution layers with kernel sizes (7,5,5,5,5),
each followed by a Batch Norm layer \cite{ioffe2015batch} and an ELU activation layer \cite{clevert2015fast}. 

We use Adam optimizer \cite{kingma2014adam} with momentum 0.9 and 0.999. Phase 1 and phase 2 share a same training schedule: the learning rate is initiated with $10^{-4}$ for the first 20 epochs and is halved every 5 epochs afterwards until stopped at the 40th epoch. We train the concept regression module separately with learning rate of $10^{-4}$ for 6 epochs. All the training process is conducted with a batch size of 256.

\paragraph{GQA}
The implementation details in the GQA setting basically follows the details on CLEVR.
To better handle the complexity in GQA,
we concatenate the object features with their corresponding bounding box coordinates to enhance the objects' location representations similar to \cite{hu2019language}. We use GloVe \cite{pennington2014glove} to initialize question word embeddings and maintain an exponential moving average with a decay rate of 0.999 to update the model parameters.

\section{Visualization of the induced concept hierarchy}
\label{app:concept_correlation}
\begin{figure}
\centering
\includegraphics[width=0.45\textwidth]{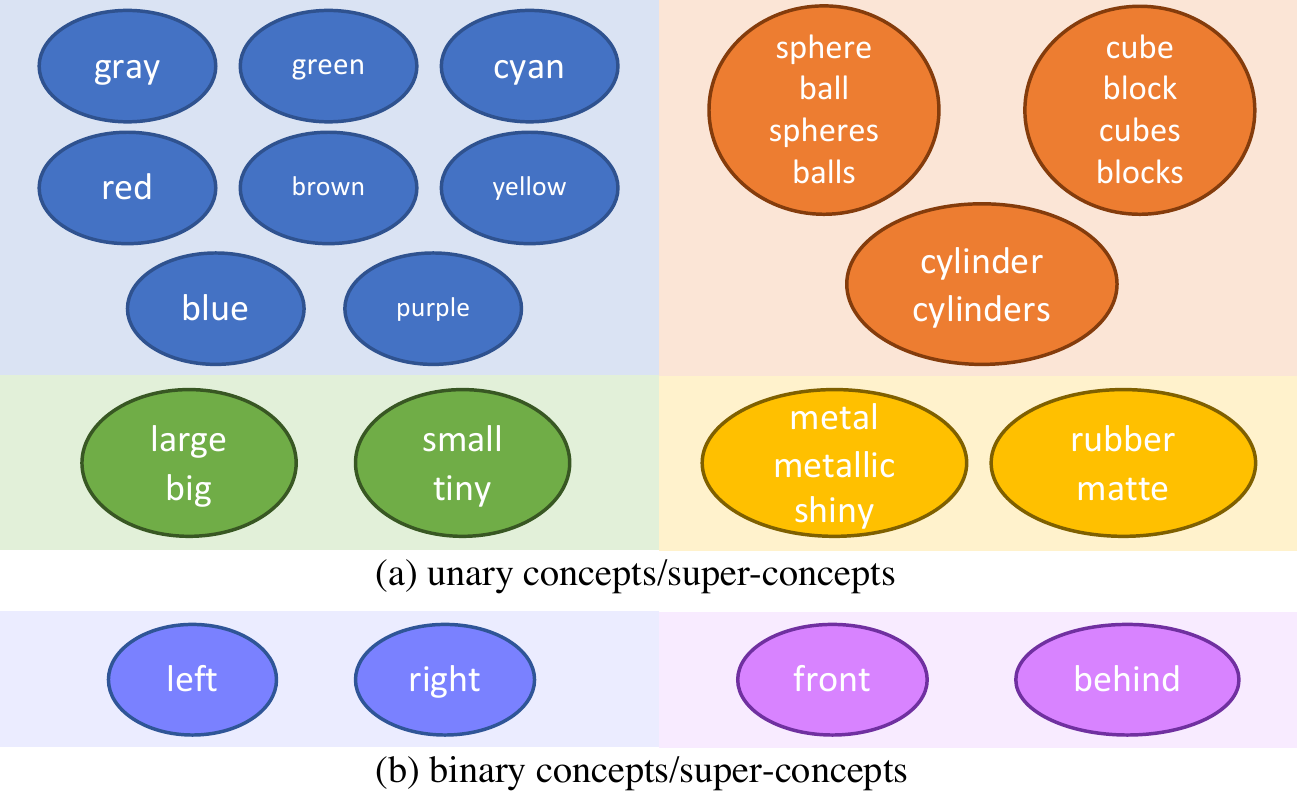}
\caption{Concepts and super concept sets. Each circle represents a concept described by the words in that circle. A super concept set comprises the concepts represented by circles of the same color.}
\label{conc_cls}
\vspace{-3mm}
\end{figure}

After visual mapping, binary coding and concept/super-concept induction, the unary concepts and super concepts are induced as shown in Figure \ref{conc_cls}; the binary concepts are 'left', 'right', 'front' and 'behind', and \{'left', 'right'\} and \{'front', 'behind'\} form two super concept sets.
The generated concept hierarchy perfectly recovers the definition in CLEVR data generator and matches human prior knowledge, showing the success of our approach.


\section{Multi-modal concept analogy}
\label{app:concept_analogy}
\begin{figure}[t]
\centering
\includegraphics[width=0.48\textwidth]{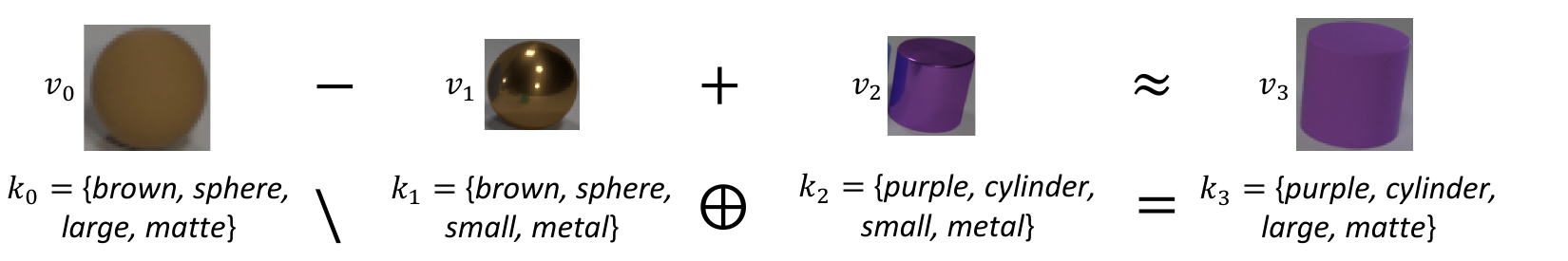}
\caption{An multi-modal analogy example enabled by our results.}
\label{ops_main}
\vspace{-1mm}
\end{figure}

Our concept induction results bridge the visual and symbolic spaces. The results enable to extend word analogy~\cite{mikolov2013distributed} (e.g., ``Madrid'' - ``Spain'' + ``France'' $\rightarrow$ ``Paris'') into the multi-modality setting.
Figure~\ref{ops_main} gives an example, starting with the initial object $v_0$ and its predicted concepts $K_0$, subtracting concepts $K_1$ and adding new concepts $K_2$ result in a new concept set $K_3$ (Figure~\ref{ops_main} (bottom)). Then if we retrieve visual object $v_i$ with each concept set $K_i$ along the path (Figure~\ref{ops_main} (top)), we have $v_0 - v_1 + v_2 \approx v_3$ in the original visual feature space.

\section{Derivation from the concept interpretation}
\label{app:concept_dis}

\label{app:visualization}
\begin{figure}[h]
\centering
\includegraphics[width=0.49\textwidth]{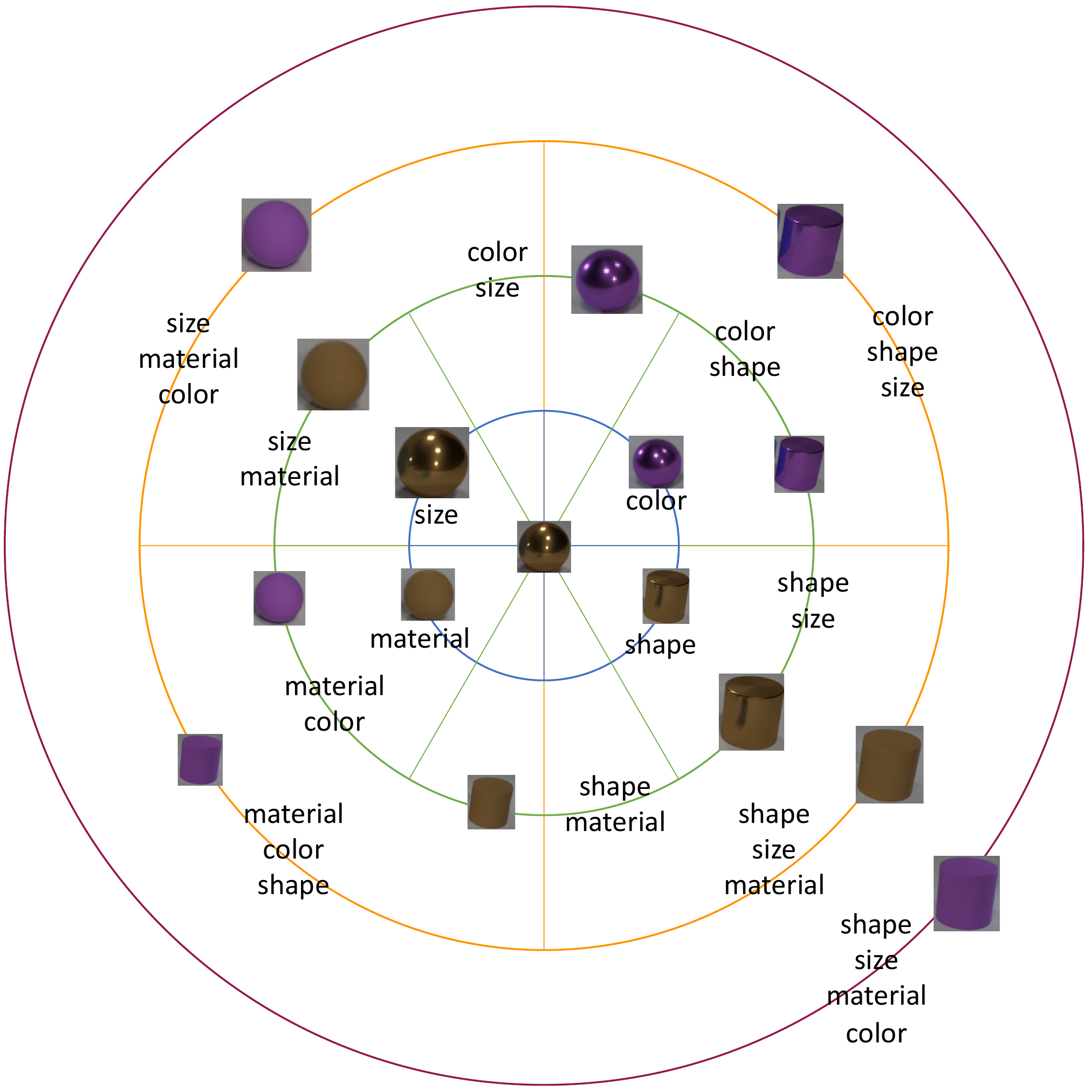}
\caption{Illustration of the semantic distance.}
\label{conc_dis}
\end{figure}

With the induced concepts and super concept sets, each object can be represented with a zero-one vector, $k$, where the entry is 1 if that object possesses the corresponding concept or 0 otherwise. Notice that the super concept sets split the whole concept set; we thereby name the entries of $k$ corresponding to one super concept set as a super concept. The super concept is thus a zero-one vector with exactly one entry to be 1. We name this pattern as the super concept constraint. Therefore, we can define the semantic distance between two visual objects by the number of different super concepts or by Eqn. (\ref{dis}). 
\begin{equation}
    \zeta_{k_1, k_2} = \frac{|k_1 \oplus k_2|^1_1}{2},
    \label{dis}
\end{equation}
where $k_1$ and $k_2$ are the concept vectors representing two objects and $\oplus$ is the operation XOR. Studying the concepts and super concept sets induced, we acknowledge that the super concept sets correspond to color, shape, size and material in semantics. Thereby, we give an example of the semantic distances of multiple objects to one object as shown in Figure \ref{conc_dis}. The circle radii indicate the semantic distances to the object at the centers of these circles. The inner three circles are segmented so that each segment represents what super concepts are different. The outer circle represents all the 4 super concepts are different between the object on that circle and the object at the center.

\begin{figure}
\centering
\includegraphics[width=0.49\textwidth]{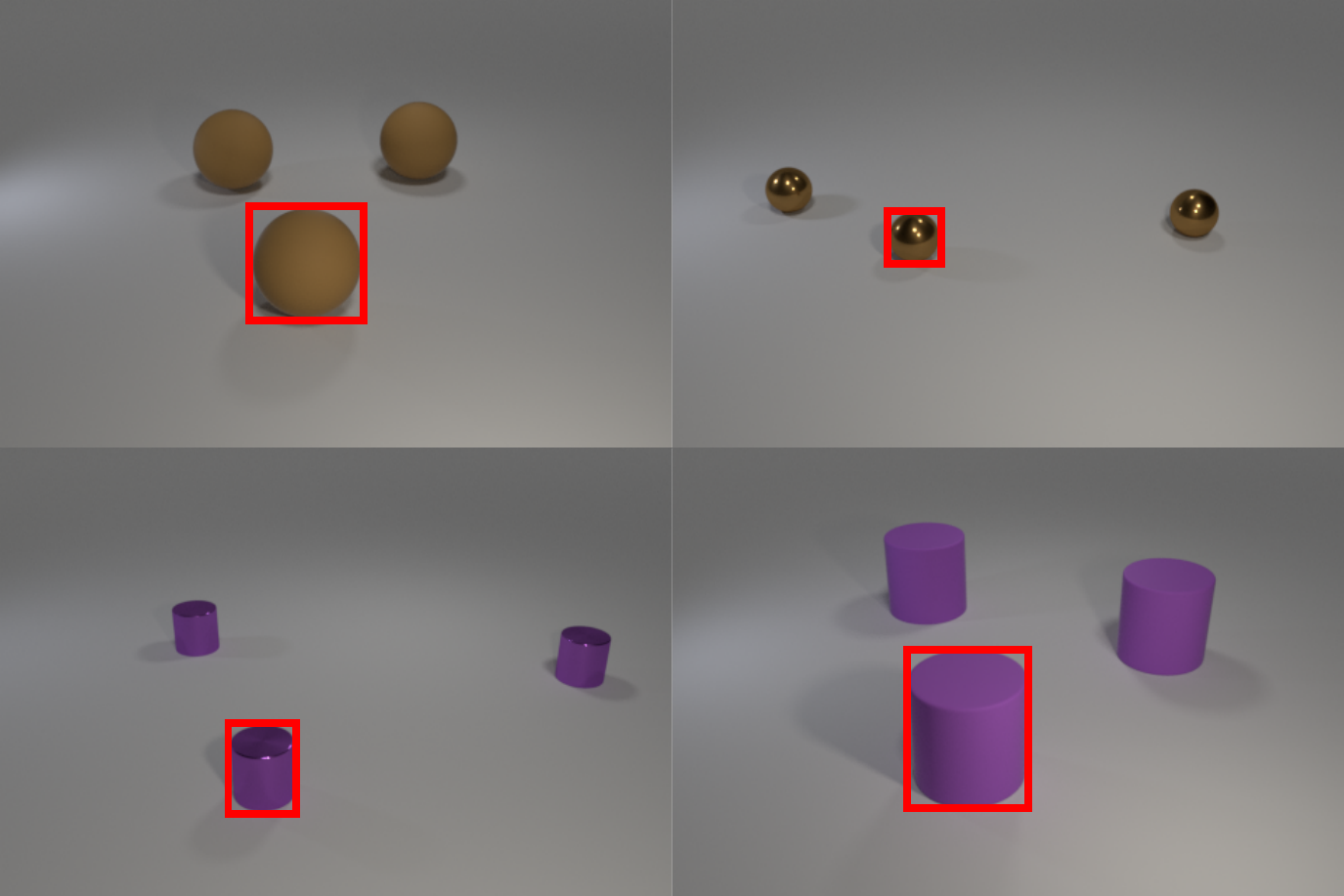}
\caption{The original images for extracting visual features. The object-level features corresponding to the objects bounded by red rectangles are used for the illustration of semantic operations in the visual feature space. }
\label{semantic_vis}
\end{figure}

\begin{figure}
\centering
\includegraphics[width=0.45\textwidth]{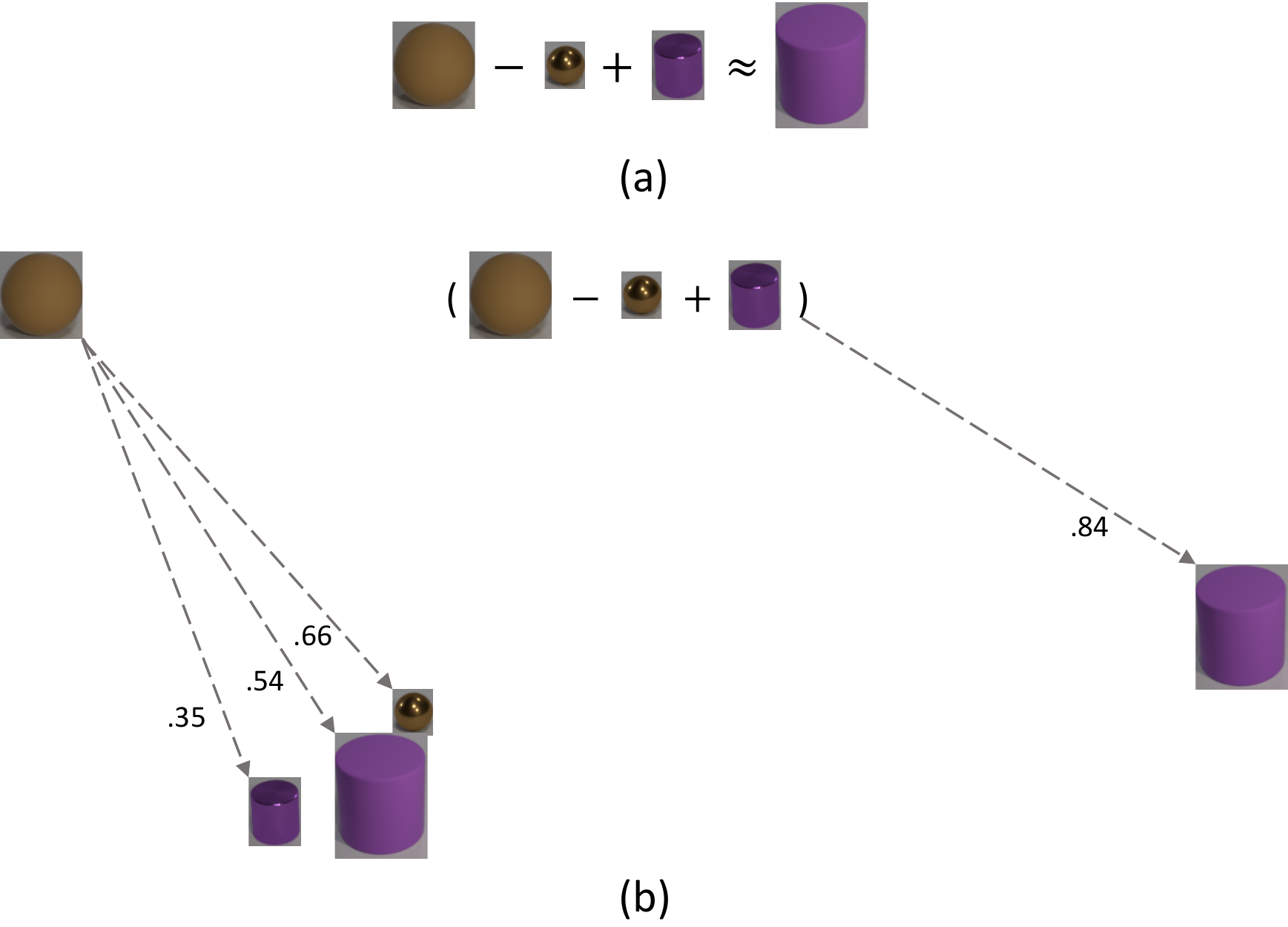}
\caption{Illustration of the semantic analogy in the visual feature space. (a) The operations on the visual features. (b) The cosine similarities between pairs of visual feature vectors. }
\label{visual_angle}
\end{figure}

\begin{figure}
\centering
\includegraphics[width=0.49\textwidth]{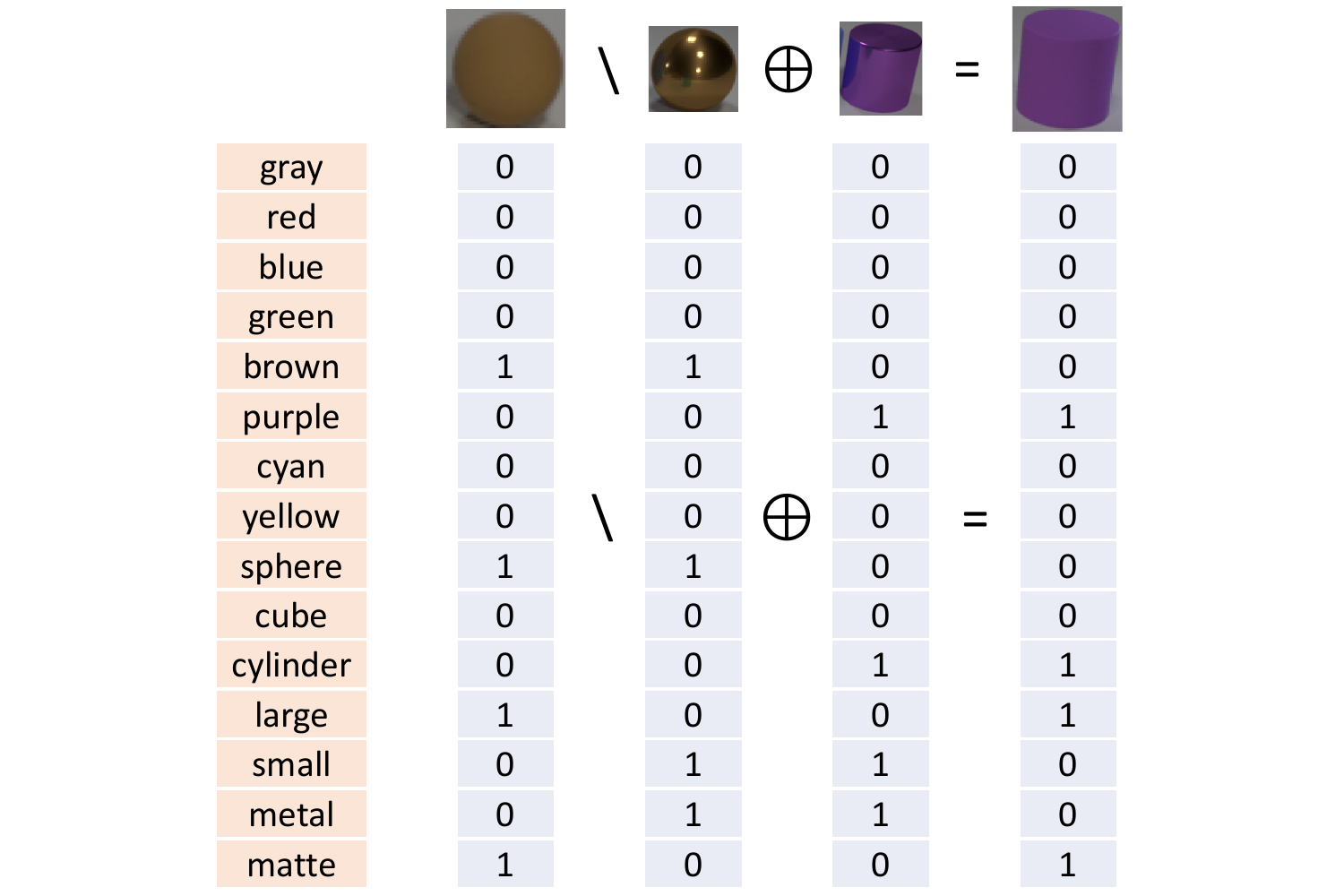}
\caption{Operations on the concept vectors.}
\label{vis_op}
\end{figure}

We can further interpret the semantic analogy in the visual feature space with the induced concept vectors. Shown in Figure \ref{semantic_vis}, we first generate four images of different objects; then, we use our trained OCCAM structure to extract the object-level features corresponding to the objects bounded by red rectangles. Shown in Figure \ref{visual_angle}(a), we can move the visual feature vector of the leftmost object closer to that of the rightmost object by subtracting and adding visual feature vectors of two other objects. The proximity between pairs of visual feature vectors is measured with cosine similarity as shown in Figure \ref{visual_angle}(b). In the concept vector space, we can define a 'minus' operation, $k_1\backslash k_2$, as eliminate the shared super concepts between $k_1$ and $k_2$ from $k_1$. We can also define a 'plus' operation, $k_1' \oplus k_2$, between a concept vector template $k_1'$ and a concept vector $k_2$ as add the super concepts of $o_2$ that $o_1'$ misses to $o_1'$. Therefore, The operations in the visual feature space can be explained with the operations we defined in the concept vector space shown in Figure \ref{vis_op}.

\section{Visualization of reasoning steps}
\label{app:vis_concept_reasoning}

\begin{figure*}
\centering
\includegraphics[width=0.85\textwidth]{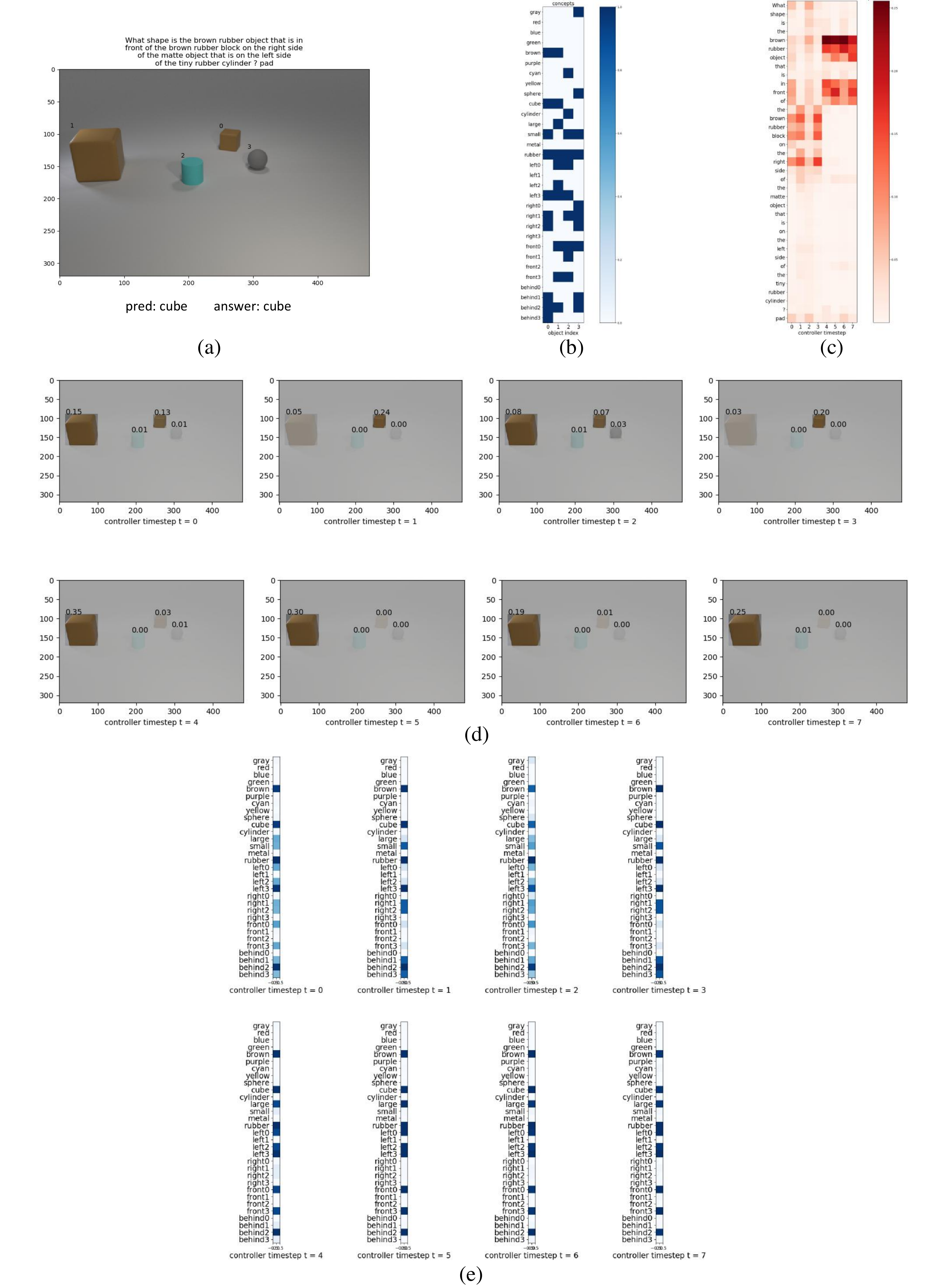}
\caption{Visualization of reasoning steps on CLEVR dataset. (a) The question, image, prediction and ground truth answer. The index of each object is shown on the upper left of the object. (b) The induced concepts of objects and relations. (c) The stepwise attentions on question words. (d) The stepwise attentions on objects. (e) The concept vector read into the memory of the reasoning module in each step.}
\label{steps}
\end{figure*}

\begin{figure*}
\centering
\includegraphics[width=0.99\textwidth]{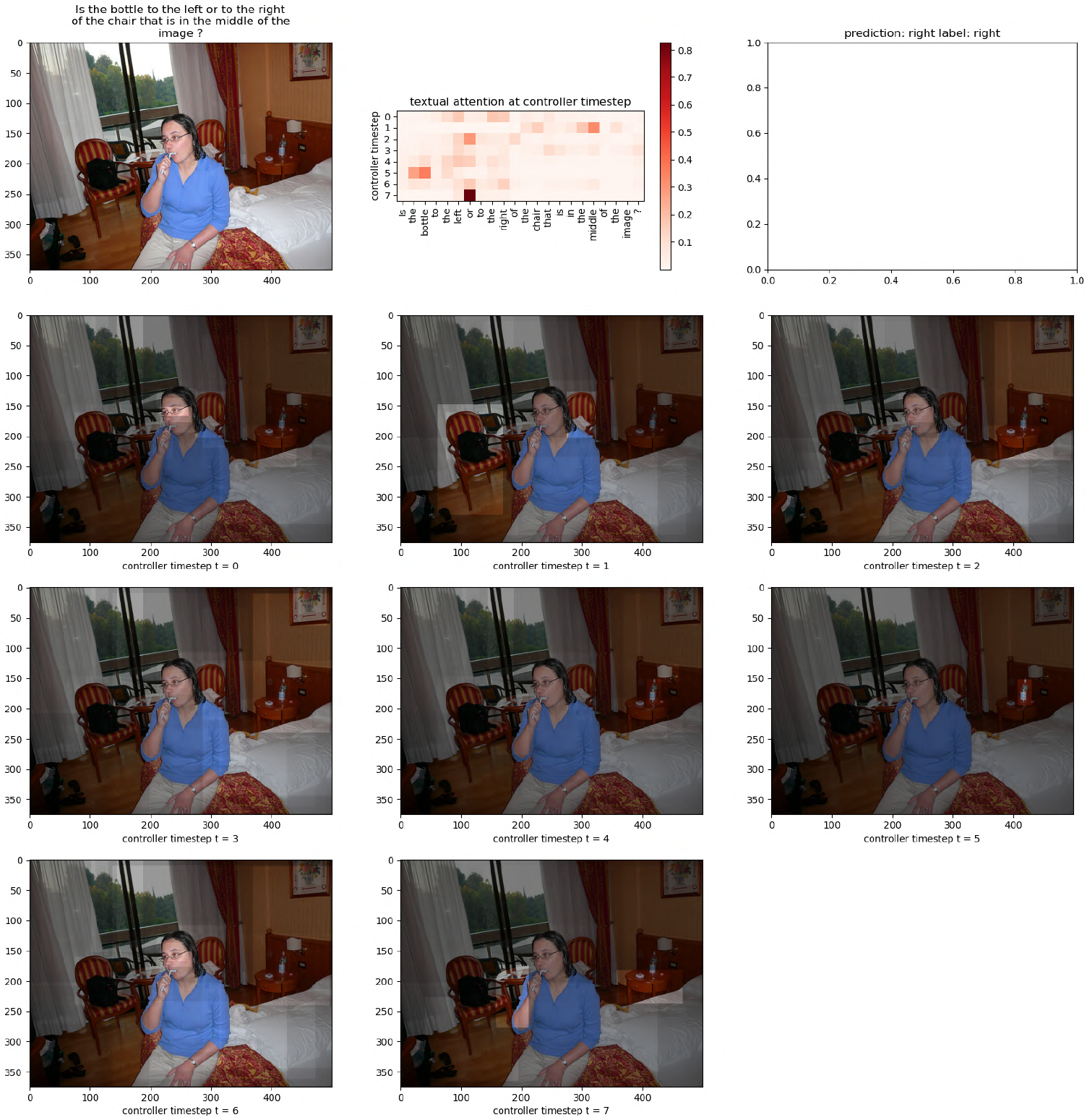}
\caption{Visualization of reasoning steps on GQA dataset. }
\label{gqa_steps}
\end{figure*}

We give an example of the compositional reasoning steps on the induced concept space of OCCAM as shown in Figure \ref{steps}. While the attention is directly imposed on the projected concept vectors in the read unit of the compositional reasoning module, the attention can be equally mapped to the concept vectors and the visual objects as the projected concept vector to the concept vector or the projected concept vector to the visual object is a one-to-one mapping relationship. We also give an example of the compositional reasoning steps on the GQA dataset shown in Figure \ref{gqa_steps}. As the dimension of the induced concept vectors is too high, here we only present the attention on objects in the image.

\section{Human study}
\label{app:human_study}
\begin{figure*}
\centering
\includegraphics[width=0.99\textwidth]{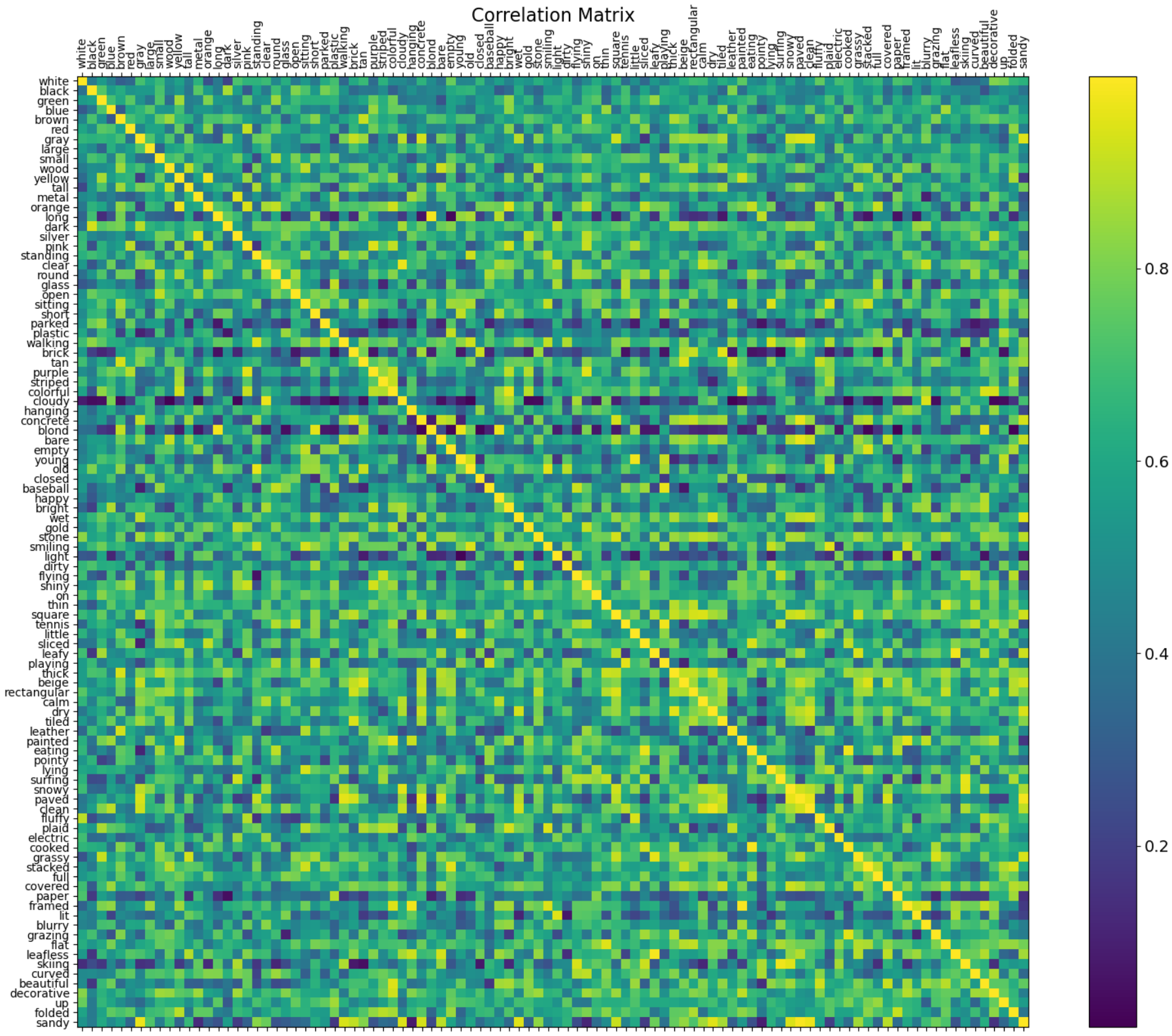}
\caption{The extended subset of GQA concept correlations.}
\label{extended_GQA_subset}
\end{figure*}

We assess the concept and super concept induction by studying how the word correlation conforms with our human knowledge. We present an extended subset of GQA concept correlations shown in Figure \ref{extended_GQA_subset}. It consists of the 98 most common single words for describing objects. Each entry in the matrix represents the conditional probability that the column attribute exists given the row attribute exists. A pair of mutual high correlation values between two words indicates that these words belong to the same concept, while the opposite means that the concepts represented by those words belong to a super concept. Therefore, we can evaluate the concept induction by assessing the conditional probabilities of synonyms or uncorrelated words for each word, because from us human understanding, a synonym is used to describe the same concept while an uncorrelated word describes a concept belonging to the same super concept. 

For each word in the extended subset words, we first let annotators choose 2 synonyms and 2 uncorrelated words from the rest 97 words. Then, rank the four chosen words in a descending order of similarity between them and the original word. Based on these annotations, we conduct two experiments: 1) measure the accuracy of classifying the chosen words to synonyms and uncorrelated words; 2) measure the Kendall tau distance \cite{kendall1938new} between the word similarity ranking based on the conditional probability and that ranking based on human knowledge. 

For the first experiment, we use a binary classifier with threshold 0.5 to classify the chosen words by humans. If a word's conditional probability given the original word is greater than the threshold, this word is classified as a synonym; if smaller, this word is classified as an uncorrelated word. The accuracy can be calculated with Eqn. (\ref{synonym_antonym_accu}).
\begin{equation}
\small
\begin{split}
    A = \frac{1}{|S|}\sum_{i \in S} \frac{1}{|W_i|}(\sum_{j \in W_i^{pos}} \mathbbm{1}(R_{i,j}>t) + \sum_{j \in W_i^{neg}} \mathbbm{1}(R_{i,j}<t)),
    \label{synonym_antonym_accu}
\end{split}
\end{equation}
where $A$ represents accuracy, $S$ is the subset of words, $W_i$ represents the set of synonyms and uncorrelated words chosen for word $i$, $R_{i,j}$ represents the conditional probability of word $j$ given word $i$ exists and $t$ is the threshold. For comparison, we also calculate the cosine similarity of word GloVe \cite{pennington2014glove} embeddings to substitute the conditional probability and serve as $R$ in Eqn. (\ref{synonym_antonym_accu}). For this setting, we tune the threshold $t$ to be 0.21 to reach the best accuracy. The result in Table \ref{gqa_human_study_accuracy} shows that our induction highly conforms with our human sense in grouping words into concepts but does not agree much with humans in grouping super concepts. By further studying specific cases, we realize that a word and its uncorrelated words defined by humans can simultaneously describe one object. For example, `white' and `black' can be used together to describe a zebra; `leafy' and `leafless' both describes a status of a plant. Such words have high correlations, which defects with our human understanding.

\begin{table}[t!]
\small
    \centering
    \caption{The accuracy of classifying synonyms and uncorrelated words. $A^{pos}$ represents the accuracy of classifying only synonyms. $A^{neg}$ represents the accuracy of classifying only antonyms. $^{\dagger}$For word2vec, we tune the threshold on ground truth, while our method is used out of the box without threshold tuning (i.e., threshold set to 0.5).}
    \begin{tabular}{c|ccc}
        \toprule
         Method & $A^{pos}$ & $A^{neg}$ & $A$\\
         \midrule
         word2vec$^{\dagger}$  & 76.02\% & 60.71\% & 68.37\% \\
         induction & 92.35\% & 63.78\% & 78.06\% \\
         \bottomrule
    \end{tabular}
    \label{gqa_human_study_accuracy}
\end{table}

The second experiment measures how the induced word proximity conforms with our human knowledge. For a word $w_i$, our annotators rank the chosen synonyms and uncorrelated words $a_i=(a_{i1}, a_{i2}, a_{i3}, a_{i4})$ with a descending order of word similarity to $w_i$ and assign a sequence of order indices $O_i^{human}=(0,1,2,3)$ to $a_i$. Then, we rank $(a_{i1}, a_{i2}, a_{i3}, a_{i4})$ with a descending order of their conditional probabilities and assign a sequence of order indices $O_i^{induce}$ to $a_i$. For comparison, we further rank $(a_{i1}, a_{i2}, a_{i3}, a_{i4})$ in a descending order of cosine similarities between the GLoVe embeddings of $(a_{i1}, a_{i2}, a_{i3}, a_{i4})$ and $w_i$ and assign a sequence of order indices $O_i^{word2vec}$ to $a_i$. The average ranking distance can be calculated with Eqn. (\ref{kendall}). 
\begin{equation}
\small
\begin{split}
    D(O^x) = \frac{1}{|S|} \sum_{i \in S} \mathcal{K}(O_i^{human},O_i^x),
    \label{kendall}
\end{split}
\end{equation}
where $D$ represents the average ranking distance, $x \in \{induce, word2vec\}$,  $\mathcal{K}$ represents the operation for calculating the normalized Kendall tau distance between two rankings. The result in Table (\ref{gqa_human_study_distance}) proves that our induction from visual language relations encodes word proximity that is more aligned with human knowledge than the one encoded by GloVe embeddings from language-only data. 

\begin{table}[t!]
\small
    \centering
    \caption{The average ranking distance to human rankings. }
    \begin{tabular}{cc}
        \toprule
         $D(O^{word2vec})$ & $D(O^{induce})$\\
         \midrule
         0.3418 & 0.2585 \\
         \bottomrule
    \end{tabular}
    \label{gqa_human_study_distance}
    \vspace{-3mm}
\end{table}

\section{Error analysis}
\begin{figure}
\centering
\includegraphics[width=0.49\textwidth]{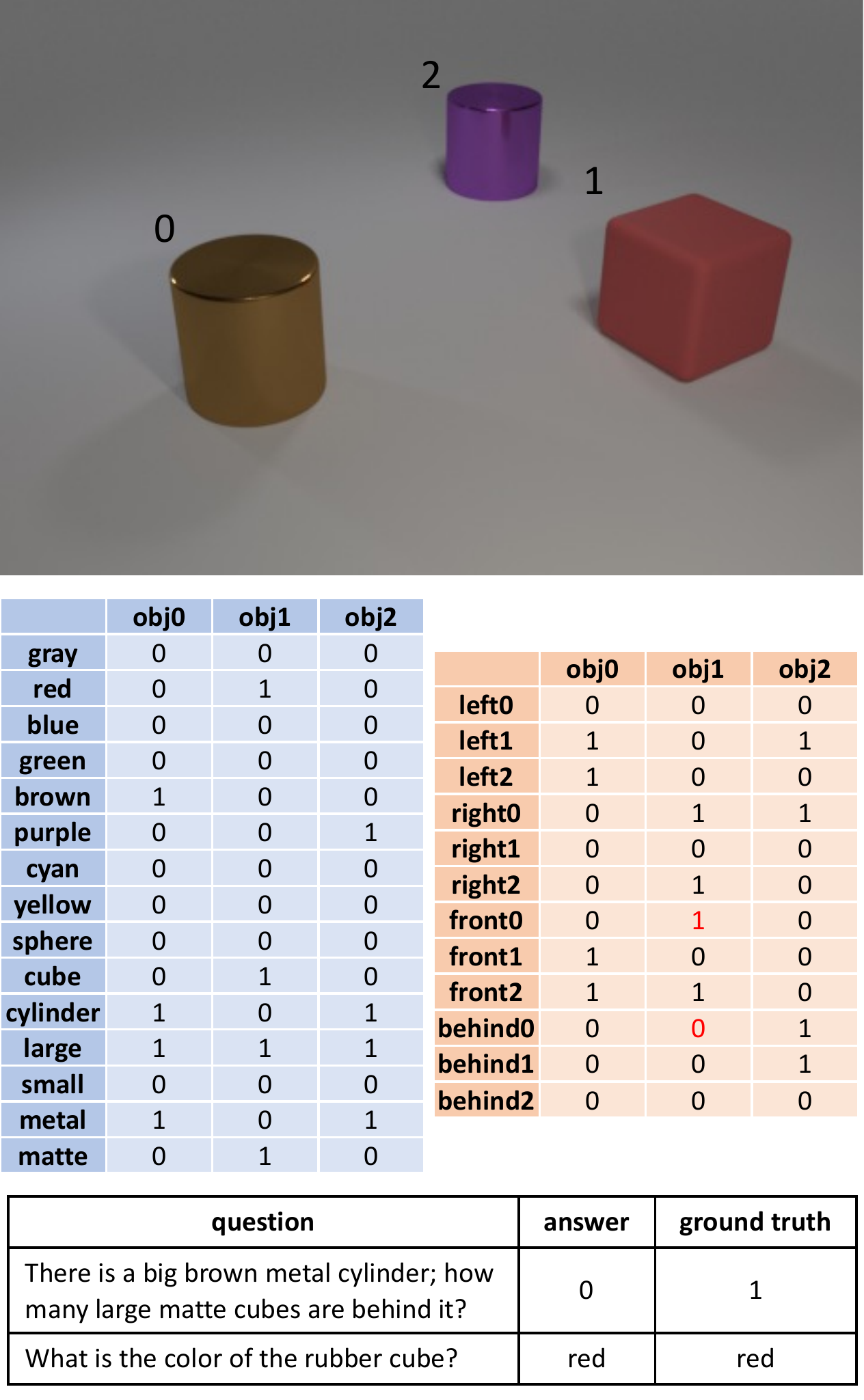}
\caption{Error analysis. The predicted unary and binary concepts corresponding to each object in the image above are shown in the tables at the middle; the digits colored in red are wrong predicted concepts. The questions, the predicted answers and the ground truth answers are shown in the table at the bottom.}
\label{error_analysis}
\end{figure}

The reasoning process may reach a false answer if 1) a concept is mentioned in the question and 2) that concept is wrongly classified for the objects ought to be attended. However, the reasoning process may still reach a correct answer if either of these two conditions is not sufficed. We present two examples in Figure \ref{error_analysis}.

\end{document}